\documentclass[sigconf]{acmart}

\usepackage{mathtools,xparse}
\usepackage{amsmath,upgreek,bm}

\usepackage[english]{babel}
\usepackage[labelformat=simple,font={small}, labelsep=period,singlelinecheck=on]{caption}
\usepackage[labelformat=simple,font={small}, labelsep=period,singlelinecheck=on]{subcaption}

\makeatletter
\def\@begintheorem#1#2{%
	\parskip 0pt 
	\item[%
	\hskip 10\p@
	\hskip \labelsep
	{{\bfseries #1\hskip 5\p@\relax#2.}}%
	]
	\it
}
\def\@opargbegintheorem#1#2#3{%
	\parskip 0pt 
	\trivlist
	\item[%
	\hskip 10\p@
	\hskip \labelsep
	{\bfseries #1\ #2\       
		\setbox\@tempboxa\hbox{(#3)}  
		\ifdim \wd\@tempboxa>\z@ 
		\hskip 5\p@\relax    
		\box\@tempboxa       
		\fi.}%
	]
	\it
}

\newtheorem{theorem}{Theorem}

\newtheorem{lemma}{Lemma}

\newtheorem{defn}{Definition}

\DeclareMathAlphabet{\pazocal}{OMS}{zplm}{m}{n}

\newcommand{\La}{\pazocal{L}}

\newcommand{\Na}{\pazocal{N}}
\newcommand{\Da}{\pazocal{D}}

\newcommand{\Xa}{\pazocal{X}}
\newcommand{\Ya}{\pazocal{Y}}
\newcommand{\BigO}{\pazocal{O}}

\def\BibTeX{{\rm B\kern-.05em{\sc i\kern-.025em b}\kern-.08emT\kern-.1667em\lower.7ex\hbox{E}\kern-.125emX}}
    
\copyrightyear{2019}
\acmYear{2019}
\setcopyright{acmlicensed}
\acmConference[In KDD '19]{The 25th ACM SIGKDD International Conference on Knowledge Discovery \& Data Mining}{August 4--8, 2019}{Anchorage, Alaska USA}
\acmBooktitle{In KDD '19: The 25th ACM SIGKDD International Conference on Knowledge Discovery \& Data Mining, August 4--8, 2019, Anchorage, Alaska USA}

\begin{document}

%
\title{Stability and Generalization     of  Graph Convolutional Neural Networks}

%

\author{Saurabh Verma}
\affiliation{%
	\institution{Department of Computer Science \\    University of Minnesota, Twin Cities}}
\email{verma076@cs.umn.edu}

\author{Zhi-Li Zhang}
\affiliation{%
	\institution{Department of Computer Science \\    University of Minnesota, Twin Cities}}
\email{zhzhang@cs.umn.edu}

%

%
\begin{abstract}
	Inspired by convolutional neural networks on 1D and 2D data, graph convolutional neural networks (GCNNs) have been developed
	for various learning tasks on graph data, and  have shown superior performance on real-world datasets. Despite their success, there is a dearth of
	{\em theoretical} explorations of GCNN models such as their generalization properties. In this paper, we take a 
	first step towards developing a deeper theoretical understanding of GCNN models  by analyzing the stability of single-layer GCNN models and deriving their  generalization guarantees in a semi-supervised graph learning setting.   In particular, we show that the algorithmic stability  of a GCNN model depends upon the largest absolute eigenvalue of its graph convolution filter. Moreover, to ensure the uniform stability needed to provide strong generalization guarantees, the  largest absolute eigenvalue must be  independent of the graph size. Our results shed new insights on the design of 
	new \& improved graph convolution filters with guaranteed algorithmic stability. We  evaluate the generalization gap and stability on various real-world graph datasets and show that the empirical results indeed support our theoretical findings. 
	To the best of our knowledge, we are the first to study stability bounds on graph learning in a semi-supervised setting and derive generalization bounds for GCNN  models.
\end{abstract}

%
%


%

%

%
\maketitle

\section{Introduction}


Building upon the huge success of deep learning in computer vision (CV) and natural language processing (NLP), 
Graph Convolutional Neural Networks (GCNNs)~\cite{kipf2016semi} have  recently been developed for tackling various learning tasks on graph-structured datasets. These models have shown superior performance on real-world datasets from various domains such as node labelling on  social networks~\cite{kipf2016variational},  link prediction in knowledge graphs~\cite{schlichtkrull2018modeling} and molecular graph classification in quantum chemistry~\cite{gilmer2017neural} . Due to the versatility of graph-structured data representation, GCNN models have been incorporated in many diverse applications, e.g., question-answer systems~\cite{song2018exploring} in NLP and/or image semantic    segmentation~\cite{qi20173d} in CV. While various versions of GCCN models have been proposed,  there is a dearth of {\em theoretical} explorations of GCNN models (\cite{xu2018powerful} is one of few exceptions which explores the {\em discriminant} power of GCNN models)---especially, in terms of their {\em generalization} properties and ({\em algorithmic}) {\em stability}. The latter is of particular import, as the stability of a learning algorithm plays a crucial role in generalization.


The  generalization of a learning algorithm can be explored in several ways.
One of the  earliest and most popular approach   is   Vapnik–Chervonenkis (VC)-theory~\cite{blumer1989learnability} which establishes generalization errors  in terms 
VC-dimensions of a learning algorithm. 
Unfortunately, VC-theory is not applicable for learning algorithms with  unbounded VC-dimensions  such as   neural networks.
Another way to show generalization is to perform the Probably Approximately Correct  (PAC)~\cite{haussler1990probably} analysis,
which is generally difficult to do in practice.  The third approach, which we adopt,  relies on deriving stability bounds of a learning algorithm, often known as {\em algorithmic stability}~\cite{bousquet2002stability}. The idea behind algorithmic stability is to understand how  the learning function changes  with small changes in the input data. 
Over the past decade, several definitions  of algorithmic stability   have  been developed~\cite{agarwal2005stability,agarwal2009generalization,bousquet2002stability,elisseeff2005stability, mukherjee2006learning},  including uniform stability, hypothesis stability, pointwise hypothesis stability, error stability and   cross-validation  stability,  each yielding either a tight or loose bound on the generalization errors.
For instance,   learning algorithm based on Tikhonov regularization satisfy the  {\em uniform stability}  criterion (the strongest stability condition among all existing forms of stability), and thus    are  generalizable.

In this paper, we take a  first step towards developing a deeper theoretical understanding of GCNN models  by analyzing the (uniform) stability of  GCNN models and thereby deriving their  generalization guarantees. For simplicity of exposition, we focus on {\em single layer}
GCNN models in a semi-supervised learning setting.  The main result of this paper is that  {\em (single layer) GCNN models with stable graph convolution filters can satisfy the strong  notion of  uniform  stability and thus are generalizable.} More specifically, 
we show that the stability  of a (single layer) GCNN model depends upon the {\em largest absolute eigenvalue} (the eigenvalue with the largest absolute value) of the graph filter it employs -- or more generally, the largest {\em singular} value if the graph filter is {\em asymmetric} -- and that the uniform stability criterion is met if the largest absolute eigenvalue (or singular value) is independent of the graph size, i.e., the  number of nodes in the graph. As a consequence of our analysis,  we establish that (appropriately) {\em normalized}    graph convolution filters such as the symmetric normalized graph Laplacian   or random walk based filters are all uniformly stable and thus are generalizable. In contrast, graph convolution filters  based on the {\em unnormalized} graph Laplacian or adjacency matrix \emph{do not} enjoy algorithmic stability, as their largest absolute eigenvalues grow as a function of the graph size. Empirical evaluations based on real world datasets support our theoretical findings:  the generalization gap and  weight parameters instability in case of unnormalized graph  filters  are \emph{significantly higher}  than those of the normalized filters. Our results shed new insights on the design of 
new \& improved graph convolution filters with guaranteed algorithmic stability. 

We remark that our GCNN generalization bounds obtained from algorithmic stability   are   non-asymptotic   in nature, i.e., they do not assume any form of data distribution. Nor do they hinge upon the complexity of the  hypothesis class,   unlike the most uniform convergence bounds. We only assume  that the activation  \& loss functions employed are Lipschitz continuous and smooth functions. These criteria are readily satisfied by several popular activation functions such as ELU (holds for $\alpha=1$), Sigmoid and/or Tanh.  
To the best of our knowledge, we are the first to study stability bounds on graph learning in a semi-supervised setting and derive generalization bounds for GCCN models. Our analysis framework remains general enough 
and can be extended to theoretical stability analyses of GCCN models beyond a semi-supervised learning setting (where there is a single and fixed  underlying graph structure) such as for the  graph classification (where there are multiple graphs). \\

\textbf{In summary, the major contributions of our paper are: }  
\begin{itemize}  
	\item   We provide the first generalization bound on single layer GCNN models based on analysis of their  algorithmic stability.
	We establish that GCNN models which employ graph filters with bounded eigenvalues  that are  independent of the graph size can satisfy the strong notion of  uniform  stability and thus are generalizable.  
	\item  Consequently, we demonstrate that many existing GCNN models that employ {\em normalized} graph filters  satisfy the strong notion of uniform stability. We also justify the importance of employing batch-normalization in  a GCNN   architecture. 
	\item   Empirical evaluations of the  generalization gap and stability using real-world datasets support our theoretical findings. 
\end{itemize}

The paper is organized as follows. Section~\ref{sec:related_work} reviews key generalization results for deep learning as well as regularized graphs and briefly discusses existing GCNN models. The main result is presented in Section~\ref{sec:model} where we introduce the needed background and establish the GCNN generalization bounds step by step. 
In Section~\ref{sec:implication}, we apply our results to  existing graph convolution filters and GCNN architecture designs. In Section~\ref{sec:exp_results}  we conduct  empirical studies which complement our theoretical analysis.
The paper is concluded in Section~\ref{sec:conclusion}   with a brief discussion of future work.

\section{Related Work}~\label{sec:related_work}

\noindent \textbf{Generalization Bounds on Deep Learning}:  
Many theoretical studies have been devoted to understanding the  representational power of neural networks by analyzing their capability as a universal function approximator as well as their depth efficiency~\cite{cohen2016convolutional, telgarsky2016benefits, eldan2016power, mhaskar2016deep, delalleau2011shallow}.  In~\cite{delalleau2011shallow} the  authors  show that the number of hidden units
in a shallow network   has to grow exponentially  (as opposed to  a linear growth in a deep network) in order to represent the same function; thus depth yields much more compact representation of a function than having a wide-breadth. It is shown in~\cite{cohen2016convolutional} that convolutional neural networks with the ReLU activation function are universal function approximators with max pooling,  but not with average pooling. The authors of~\cite{neyshabur2017exploring} authors explore which complexity measure is more appropriate for explaining the generalization power of deep learning. The work most closest to ours is \cite{hardt2015train} where the authors derive  upper bounds on the generalization errors for stochastic gradient methods. While also utilizing the notion of uniform stability~\cite{bousquet2002stability}, their analysis is concerned with the impact of  SGD learning rates. More recently, through empirically evaluations on    real-world  datasets, it has been argued in~\cite{zhang2016understanding} that the traditional measures of model complexity are not sufficient to explain the generalization ability  of neural networks. Likely, in~\cite{kawaguchi2017generalization} several open-ended questions are posed regarding the (yet unexplained) generalization capability of neural networks, despite  their possible algorithmic instability,  non-robustness, and sharp minima.  \\

\noindent \textbf{Generalization Bounds on Regularized Graphs}:  Another line of work concerns with generalization bounds on regularized graphs   in transductive settings~\cite{belkin2004regularization, cortes2008stability, ando2007learning, sun2014manifold}. 
Of the most interest to ours is~\cite{belkin2004regularization} where the authors provide theoretical guarantees for the generalization error based on Laplacian regularization, which are also derived based on the notion of algorithmic stability.   Their generalization  estimate is \emph{inversely proportional} to the second smallest eigenvalue of the graph Laplacian. Unfortunately this estimate may be  not yield desirable guarantee  as the second smallest eigenvalue is dependent on both the graph structure and its size; it is in general difficult to remove this dependency via normalization. In contrast, our estimates are \emph{directly proportional} to  the  largest absolute eigenvalue (or the largest singular value of an asymmetric graph filter), and can easily be made independent of the graph size by performing appropriate   Laplacian normalization. \\

\noindent \textbf{Graph Convolution Neural Networks}:  Coming from graph signal processing~\cite{shuman2013emerging} domain, GCNN is defined as the  problem of  learning filter parameters in the graph Fourier transform~\cite{bruna2013spectral}.  Since then rapid progress has been made and GCNN model have improved in many aspects~\cite{kipf2016semi, atwood2016diffusion, li2018adaptive, duvenaud2015convolutional,puy2017unifying, dernbach2018quantum, verma2018graph}. For instance in~\cite{li2018adaptive} parameterize graph filters using residual Laplacian matrix and in~\cite{such2017robust} authors used simply polynomial of adjacency matrix.   Random walk and  quantum walk based graph convolutions are also been proposed recently~\cite{puy2017unifying, dernbach2018quantum, zhang2019quantum}. Similarly,   graph convolutional operation has been generalized with the graph capsule notion in~\cite{verma2018graph}. The authors of~\cite{hamilton2017inductive,velivckovic2018deep}  have also applied  graph convolution to large graphs.   Message passing neural networks (MPNNs) are also been developed~\cite{lei2017deriving, gilmer2017neural,dai2016discriminative, garcia2017learning} which can be viewed as GCNN model since the  notion of    graph convolution operation   remains the same.  MPNNs can also be break into two step process where edge features are updated though message passing and then node features are updates using the information encoded in its nearby edges. This is similar to Embedding belief propagation message passing algorithm proposed in~\cite{dai2016discriminative}.  Several attempts have also been made to convert graph into regular grid structure for straight forwardly applying standard 2D or 1D CNNs~\cite{niepert2016learning, tixier2018graph}. A very  tangential approach  was taken in~\cite{kondor2018covariant} where authors design covariant neural network based on group theory for computing   graph representation.

\section{Stability and Generalization Guarantees For GCNNs}\label{sec:model}
To derive generalization guarantees of GCNNs based on algorithmic  stability analysis, we adopt the  strategy devised in~\cite{bousquet2002stability}. It relies on bounding the output difference of a loss function due to a single data point perturbation. As stated earlier,  there exist several different notions of algorithmic stability~\cite{bousquet2002stability,mukherjee2006learning}.  In this paper, we focus on the strong notion of  {\em uniform stability} (see Definition~\ref{def:random_uniform_stability}). 

\subsection{Graph Convolution Neural Networks}
\noindent \textbf{Notations}: Let   $G=(V,E,\mathbf{A})$ be a graph   where $V$ is the vertex set, $E$ the edge set and $\mathbf{A}$ the       adjacency matrix, with  $N=|V|$ the graph size. We define the standard graph Laplacian $\mathbf{L}  \in \mathbb{R}^{N \times N} $ as $\mathbf{L}=\mathbf{D}-\mathbf{A} $, where $\mathbf{D}$ is the degree matrix. We define a graph filter, $g(\mathbf{L})  \in \mathbb{R}^{N \times N}$ as  a    function of   the graph Laplacian $\mathbf{L}$  or a normalized (using $\mathbf{D}$) version of it. Let $\mathbf{U} \mathbf{\Lambda}\mathbf{U}^T$ be the eigen decomposition of $L$, with $\mathbf{\Lambda}=diag[\lambda_i]$ the diagonal matrix of $\mathbf{L}$'s eigenvalues.  Then $g(\mathbf{L})=\mathbf{U}g(\mathbf{\Lambda})\mathbf{U}^T$, and its eigenvalues $\lambda^{(g)}_i=\{g(\lambda_i),  1\leq  i \leq N\}$. We define $\lambda^{\max}_G =\max_i\{|\lambda^{(g)}_i|\}$, referred to as the {\em largest absolute eigenvalue}\footnote{This definition is valid for a symmetric graph filter $g(\mathbf{L})$, or the matrix is normal. More generally, 
	$\lambda^{\max}_G$ is defined as the largest singular value of $g(\mathbf{L})$.} of the graph filter $g(\mathbf{L})$. Let $m$ is the number of training samples depending on $N$ as $m\leq N$.


Let $\mathbf{X}\in \mathbb{R}^{N \times D}$ be a node feature matrix ($D$ is the input dimension) and $ \bm{\uptheta} \in  \mathbb{R}^{D}$ be the learning parameters. With a slight abuse of notation, we will represent both a node (index) in a graph $G$ and its feature values by   $ \mathbf{x} \in \mathbb{R}^{D} $.  $\Na(\mathbf{x})$ denotes a set of the neighbor indices at most $1-$hop distance away from  node $\mathbf{x}$ (including $ \mathbf{x}$). 
Here the $1-$hop distance neighbors are determined using the $g(\mathbf{L})$ filter matrix. Finally, $G_{\mathbf{x}}$ represents the ego-graph extracted at node $\mathbf{x}$ from $G$. \\ 

\vspace*{1pt}
\noindent \textbf{Single Layer GCNN (Full Graph View)}: Output function of a single layer GCNN model -- on all graph nodes together -- can be written in a compact matrix form as follows,
\begin{equation}\label{eq:gcnn}
\begin{split}
f(\mathbf{X}, \bm{\uptheta}) & = \sigma\Big( g(\mathbf{L})\mathbf{X} \bm{\uptheta} \Big) \\
\end{split}
\end{equation}
where $g(\mathbf{L})$ is a graph filter. Some commonly used graph filters are 
a  linear function of $\mathbf{A}$ as $g(\mathbf{L}) = \mathbf{A} + \mathbf{I}$~\cite{xu2018powerful} (here $\mathbf{I}$ is the identity matrix)  or a Chebyshev  polynomial of $\mathbf{L}$~\cite{defferrard2016convolutional}.  \\

\noindent \textbf{Single Layer GCNN (Ego-Graph View)}: We will work with the notion of  ego-graph for each node (extracted from $G$)  as it contains the \emph{complete} information needed  for computing the output of a single layer GCNN model. We can re-write the Equation~(\ref{eq:gcnn}) for a single node prediction as,  
\begin{equation}\label{eq:gcnn_node}
\begin{split}
f(\mathbf{x},\bm{\uptheta} ) &= \sigma\Big( \sum_{\mathclap{\substack{j\in \\ \Na(\mathbf{x})}}}  e_{\cdot j}  \mathbf{x}^{T}_{j}  \bm{\uptheta} \Big)  \\
\end{split}
\end{equation}
where $e_{\cdot j} \in \mathbb{R} =[g(\mathbf{L})]_{\cdot j}$ is the  weighted edge (value)   between node $\mathbf{x}$ and its neighbor $\mathbf{x}_j$, $j\in \Na(\mathbf{x})$ if and only $e_{\cdot j} \neq 0$. The size of an ego-graph depends upon $g(\mathbf{L})$. We   assume that the filters are localized to the $1-$hop neighbors, but  our analysis is applicable to $k-$hop neighbors. For further notational clarity, we will consider the case $D=1$, and thus $f(\mathbf{x},\bm{\uptheta}_S ) = \sigma\Big( \sum_{j\in \\ \Na(\mathbf{x})}  e_{\cdot j}  \mathbf{x}_{j}  \bm{\uptheta}_S \Big)$.  Our analysis holds  for the  general $D-$dimensional case. 

\subsection{Main Result}
The main result of the paper is stated in Theorem~\ref{thm:gcnn_gen_bound}, which provides a bound on the generalization gap for single layer GCNN models. This gap is defined as the difference between  the  generalization error $R(\cdot)$ and empirical error $R_{emp}(\cdot)$ (see definitions in Section~\ref{sec:prelim}).

\begin{theorem}[\textbf{GCNN Generalization Gap}]\label{thm:gcnn_gen_bound} \textit{Let $A_S$ be a  single layer GCNN model   equipped with the graph convolution filter $g(L)$,  and trained on a dataset  $S$ using  the SGD algorithm for $T$ iterations. Let the loss \& activation functions be Lipschitz-continuous and smooth. Then the  following expected generalization gap   holds with probability at least  $1 -\delta $,  with $\delta \in (0,1)  $,}
	\begin{equation*} 	
	\begin{split}
	\mathbf{E}_{\textsc{sgd}}[R(A_S) - R_{emp}(A_S)] & \leq  \frac{1}{m}\BigO \big( (\lambda_{G}^{\max})^{2T}\big) +  \\
	& \hspace{1.5em} \Big( \BigO \big((\lambda_{G}^{\max})^{2T} \big)+M\Big)\sqrt{\frac{\log \frac{1}{\delta}}{2m}} \\
	\end{split}	 
	\end{equation*} 	
	where the expectation $\mathbf{E}_{\textsc{sgd}}$ is taken over   the randomness inherent in SGD, $m$ is the number of training samples and $M$  a constant depending on the loss function.
	
\end{theorem} 

\noindent \textbf{Remarks}: Theorem~\ref{thm:gcnn_gen_bound}       establishes a key  connection between the      generalization gap and  the graph    filter eigenvalues. A GCNN model is uniformly stable if  the bound  converges to zero as $m \rightarrow \infty$.
In particular, we see that if  $\lambda_{G}^{\max}$ is independent of the graph size, the generalization gap decays at the rate of $\BigO(\frac{1}{\sqrt{m}})$, yielding the tightest bound possible. Theorem~\ref{thm:gcnn_gen_bound} sheds light on the design of  stable graph filters with generalization guarantees. \\


\noindent  \textbf{Proof Strategy}: 
We need to tackle several technical challenges in order to  obtain  the generalization  bound in Theorem~\ref{thm:gcnn_gen_bound}.
\begin{enumerate} 
	\item \textbf{Analyzing GCNN Stability w.r.t. Graph Convolution}: We analyze the stability of a graph convolution function  under the single data perturbation. For this purpose, we   separately  bound the  difference on weight parameters from the graph convolution operation in  the GCNN output function. 
	\item \textbf{Analyzing GCNN Stability w.r.t. SGD algorithm}:  GCNNs  employ the randomized stochastic gradient descent algorithm (SGD) for optimizing the   weight parameters. Thus,  we need to bound the difference in the expected value over the learned weight parameters  under single data perturbation and establish stability bounds. For this, we analyze the uniform stability of SGD in the context of GCNNs. We adopt the same strategy as in~\cite{hardt2015train} to obtain uniform stability of GCNN models, but with fewer      assumptions compared with the general case~\cite{hardt2015train}.
\end{enumerate}	

\subsection{Preliminaries}~\label{sec:prelim}
\noindent \textbf{Basic Setup}: Let $\Xa$ and  $\Ya$ be a  a subset of a Hilbert space and define $Z = \Xa \times \Ya $. We define $\Xa$  as the input space and $\Ya$ as the output space. Let   $\mathbf{x} \in \Xa , \mathbf{y} \in \Ya \subset R $ and $S$ be a training set $S=\{\mathbf{z}_1=(\mathbf{x}_1,y_1),\mathbf{z}_2=(\mathbf{x}_2,y_2),...,\mathbf{z}_m=(\mathbf{x}_m,y_m)\}$. We introduce two more notations below:

Removing  $i^{th}$  data point in the set  $S$ is represented as,

$$S^{\backslash i}=\{\mathbf{z}_1,....,\mathbf{z}_{i-1},\mathbf{z}_{i+1},.....,\mathbf{z}_m \}$$

Replacing the $i^{th}$ data point in $S$ by $\mathbf{z}_{i}^{'}$ is represented as,

$$S^{i}=\{\mathbf{z}_1,....,\mathbf{z}_{i-1},\mathbf{z}_{i}^{'},\mathbf{z}_{i+1},.....,\mathbf{z}_m \}$$ \\
\noindent \textbf{General Data Sampling Process}: Let $\Da$ denote an unknown distribution from which $\{\mathbf{z}_1,....,\mathbf{z}_m\}$ data points are sampled to form a training set $S$. Throughout the paper, we assume all samples (including the replacement sample) are i.i.d. unless mentioned otherwise. Let $\mathbf{E}_S[f]$  denote the expectation   of the function $f$ when  $m$ samples are drawn from $\Da$ to form the training set $S$. Likewise, let $\mathbf{E}_z[f]$ denote the expectation of the function $f$ when  $\mathbf{z}$ is sampled  according to $\Da$. \\

\noindent \textbf{Graph Node Sampling Process}:  At first it may  not be clear on how to describe the sampling procedure of nodes from  a graph $G$ in the context of GCNNs for performing semi-supervised learning. For our purpose, we consider {\em ego-graphs} formed by the $1-$hops neighbors  at each node as a single data point. This ego-graph is \emph{necessary and sufficient} to compute the single layer GCNN output as shown in Equation~(\ref{eq:gcnn_node}). We assume node data points are sampled in an i.i.d. fashion by first choosing a node $\mathbf{x}$   and then extracting its neighbors from $G$ to form an ego-graph.  \\

\noindent \textbf{Generalization Error}: Let $A_S $ be a  learning algorithm trained on dataset $S$.  $A_S $ is defined as a function from $Z^{m}$ to $(\Ya)^{X}$. 
For GCNNs, we set $A_S =f(\mathbf{x},\bm{\uptheta}_S )$. Then generalization error or risk $R(A_S)$ with respect to a  loss function $\ell:\mathbf{Z}^{m} \times \mathbf{Z}  \rightarrow \mathbb{R}$ is defined as, 
$$R(A_S):=\mathbf{E}_z[\ell(A_S,\mathbf{z})]=\int \ell(A_S,\mathbf{z}) p(\mathbf{z}) d\mathbf{z}.$$

\noindent \textbf{Empirical Error}: Empirical risk $R_{emp}(A_S)$ is defined as,
$$R_{emp}(A_S):=\frac{1}{m}\sum\limits_{j=1}^{m}\ell(A_S,\mathbf{z}_j).$$

\noindent \textbf{Generalization Gap}: When $A_S$ is a randomized algorithm,  we consider the expected generalization  gap as shown below,
$$\epsilon_{\textrm{gen}}:=\mathbf{E}_{A}[R(A_S) - R_{emp}(A_S)].$$
Here the expectation $\mathbf{E}_{A}$ is taken over the inherent randomness   of  $A_S$. For instance, most learning algorithms employ Stochastic Gradient descent (SGD) to learn the weight parameters. SGD introduces randomness due to the random order it uses to choose samples for batch processing. In our analysis, we only consider randomness in $A_S$  due to SGD and ignore the randomness introduced by parameter  initialization. 
Hence, we  will replace $\mathbf{E}_{A}$ with $\mathbf{E}_{\textsc{sgd}}$.   \\

%

\noindent \textbf{Uniform Stability of Randomized Algorithm}: For a randomized algorithm,  uniform stability is defined as follows,
\begin{defn}[\textbf{Uniform Stability}]\label{def:random_uniform_stability}
	\textit{A randomized learning algorithm $A_S$ is $\beta_m-$uniformly  stable with respect to a loss function $\ell$,  if  it satisfies, }
	$$ \underset{S,z}{\sup}|\mathbf{E}_{A}[\ell(A_{S},\mathbf{z})] -\mathbf{E}_{A}[\ell(A_{S^{\backslash i}},\mathbf{z})]|  \leq  \beta_m $$
\end{defn}
For our convenience, we will work with the following definition  of uniform stability, 
$$   \underset{S,z}{\sup}|\mathbf{E}_{A}[\ell(A_{S},\mathbf{z})] -\mathbf{E}_{A}[\ell(A_{S^{  i}},\mathbf{z})]|\leq 2\beta_m   $$
which follows immediately from the fact that,
\begin{equation*}
\begin{split}
&  \underset{S,z}{\sup}|\mathbf{E}_{A}[\ell(A_{S},\mathbf{z})] - \mathbf{E}_{A}[\ell(A_{S^{  i}},\mathbf{z})]| \leq  \Big(\underset{S,z}{\sup}|\mathbf{E}_{A}[\ell(A_{S},\mathbf{z})] - \\ &  \mathbf{E}_{A}[\ell(A_{S^{ \backslash i}},\mathbf{z})]|\Big) + \Big(   
\underset{S,z}{\sup}|\mathbf{E}_{A}[\ell(A_{S^{i}},\mathbf{z})] -\mathbf{E}_{A}[\ell(A_{S^{ \backslash  i}},\mathbf{z})]| \Big) \\
\end{split}
\end{equation*} 

\noindent \textbf{Remarks}: Uniform stability imposes an upper bound on the difference in losses due to a removal (or change) of a single data point from the set (of size $m$) for all possible combinations of $S,z$. Here, $\beta_{m}$ is a function     of $m$ (the number of training samples). Note that there is a subtle difference between Definition~\ref{def:random_uniform_stability} above and the uniform stability of randomized algorithms defined in~\cite{elisseeff2005stability} (see Definition $13$ in~\cite{elisseeff2005stability}). The authors in~\cite{elisseeff2005stability}  are concerned with random elements associated with the cost function such as those induced by bootstrapping, bagging or  initialization process. However, we focus on the randomness due to the learning procedure, i.e., SGD. \\

\noindent \textbf{Stability Guarantees}: A randomized learning algorithm with uniform stability yields the following bound on   generalization gap:
\begin{theorem}[\textbf{Stability Guarantees}]\label{thm:generlization_error} \textit{A uniform stable randomized algorithm $(A_S,\beta_m)$ with a bounded loss function $0\leq \ell(A_S,\mathbf{z}) \leq M$, satisfies following generalization bound with probability at-least $1-\delta$,  over the random draw of $S$,$\mathbf{z}$ with $\delta \in (0,1)$,  }
	$$\mathbf{E}_{A}[R(A_S) - R_{emp}(A_S)] \leq 2\beta_{m}+\big(4m\beta_{m}+M\big)\sqrt{\frac{\log \frac{1}{\delta}}{2m}}.$$ 
\end{theorem}
\noindent \textbf{Proof}: The proof for Theorem~\ref{thm:generlization_error} mirrors that of Theorem 12 (shown in~\cite{bousquet2002stability} for {\em deterministic} learning algorithms). For the sake of completeness, we include the proof in Appendix based on our definition of uniform stability $ :=   \underset{S,z}{\sup}|\mathbf{E}_{A}[\ell(A_{S},\mathbf{z})] -\mathbf{E}_{A}[\ell(A_{S^{  i}},\mathbf{z})]|\leq 2\beta_m   $. \\

\noindent \textbf{Remarks}: The generalization bound is meaningful if   
the bound converges to 0 as $m \rightarrow \infty$.  This occurs  when $\beta_{m}$  decays faster than
$\BigO(\frac{1}{\sqrt{m}})$; otherwise  the generalization gap does not approach to zero as $m\rightarrow \infty$.  Furthermore, generalization gap produces tightest bounds when $\beta_{m}$   decays at $\BigO(\frac{1}{m})$ which   is the most stable state possible for a learning algorithm. \\

\noindent \textbf{$\sigma-$Lipschitz  Continuous and Smooth Activation Function}: Our bounds hold for all activation functions which are Lipschitz-continuous and smooth.  An  activation function $\sigma(x)$ is Lipschitz-continuous if $|\nabla \sigma(x)| \leq   \alpha_{\sigma}$, or equivalently,  $|\sigma(x)-\sigma(y)| \leq \alpha_{\sigma}|x-y|$.
We further require $\sigma(x)$ to be smooth, namely,  $|\nabla \sigma(x) -\nabla \sigma(y)| \leq \nu_{\sigma} |x-y|$. This assumption  is more strict but  necessary for establishing the strong notion of uniform stability. Some common activation functions satisfying the above conditions are ELU (with $\alpha=1$), Sigmoid, and Tanh. 

\noindent \textbf{$\ell-$Lipschitz Continuous and Smooth Loss Function}: We  also  assume that the loss function  is Lipschitz-continuous and smooth,
\begin{equation*} 	
\begin{split}
\big|\ell\big(f(\cdot), y\big) -\ell\big(f^{'}(\cdot),y\big) \big| & \leq \alpha_{\ell} \big| f (\cdot) - f^{'}(\cdot) ) \big|,   \\
\mbox{and }\big|\nabla \ell\big(f(\cdot), y\big) -\nabla\ell\big(f^{'}(\cdot)  ,y\big) \big| & \leq \nu_{\ell} \big| \nabla f(\cdot) - \nabla f^{'}(\cdot) \big|.   \\
\end{split}
\end{equation*}
Unlike in~\cite{hardt2015train}, we define Lipschitz-continuity with respect to the function argument rather than the weight parameters, a relatively weak assumption.

\subsection{Uniform Stability of GCNN Models}
The crux of our main result relies on  showing  that GCNN models are uniformly stable as stated in Theorem~\ref{thm:gcnn-uniform-stability} below.

\begin{theorem}[\textbf{GCNN Uniform Stability}]\label{thm:gcnn-uniform-stability} \textit{ Let the loss \& activation be  Lipschitz-continuous and smooth functions. Then a single layer GCNN model trained using the SGD algorithm for $T$ iterations is   $\beta_{m}-$uniformly stable, where}
	$$ \beta_{m}  \leq   \Big( \eta \alpha_{\ell}  \alpha_{\sigma}    \nu_{\ell}  (\lambda_{G}^{\max})^2 \sum_{t=1}^{T} \big( 1 + \eta\nu_{\ell}  \nu_{\sigma}  (\lambda_{G}^{\max})^2 \big)^{t-1}\Big) /m.    $$  
\end{theorem}
\noindent \textbf{Remarks}: Plugging the bound on $\beta_m$ in Theorem~\ref{thm:generlization_error} yields the main result of our paper.

Before we proceed to prove this theorem, we first explain what is meant by training a single layer GCNN using SGD on datasets $S$ and $S^{i}$ which differ in one data point, following the same line of reasoning as in~\cite{hardt2015train}. Let $Z=\{\mathbf{z}_1,\ldots,\mathbf{z}_t,\ldots,\mathbf{z}_T\}$ be a sequence of samples, where $\mathbf{z}_t$ is an i.i.d. sample drawn from $S$ at the $t^{th}$ iteration of SGD during a training run of the GCCN\footnote{ One way to generate the sample sequence is to choose a node index $i_t$  uniformly at random   from the set $\{1,\ldots,m\}$ at each step $t$.  Alternatively, one can first choose a random permutation of $\{1, \dots, m\}$ and then process the samples accordingly.  Our analysis holds for both cases.}. 
Training the same GCCN using SGD on $S^{i}$ means that we supply the same sample sequence to the GCCN except that if $\mathbf{z}_t=(\mathbf{x}_i,y_i)$ for some $t$ ($1\leq t \leq T$), we replace it with $\mathbf{z}^{'}_t=(\mathbf{x}'_i,y'_i)$, where $i$ is the (node) index at which $S$ and $S^{i}$ differ. We denote this sample sequence by $Z'$.  Let $\{\bm{\uptheta}_{S,0}$ ,  $\bm{\uptheta}_{S,1}$  $,\dots,$ $\bm{\uptheta}_{S,T}\}$ and  $\{\bm{\uptheta}_{S^i,0},$  $\bm{\uptheta}_{S^i,1}$ $,\dots,$  $\bm{\uptheta}_{S^i,T} \}$ denote the corresponding sequences of the weight parameters learned by running SGD on $S$ and $S^{i}$, respectively. 
Since the parameter initialization is kept same, $\bm{\uptheta}_{S,0} = \bm{\uptheta}_{S^i,0}$. In addition, if $k$ is the first time that the sample sequences $Z$ and $Z'$ differ, then $\bm{\uptheta}_{S,t} = \bm{\uptheta}_{S^i,t}$ at each step $t$ before $k$, and at the $k^{th}$ and subsequent steps, $\bm{\uptheta}_{S,t}$ and $\bm{\uptheta}_{S^i,t}$ diverge. The key in establishing the uniform stability of a GCNN model is to bound the difference in losses when training the GCNN using SGD on $S$ {\em vs.} $S^{i}$.  As stated earlier in the proof strategy, we proceed in two steps. \\

\noindent \textbf{Proof Part I (Single Layer GCNN   Bound)}: We first bound the expected loss by separating the factors due to the graph convolution operation {\em vs.} the expected difference in the filter weight parameters learned via SGD on two datasets $S$ and $S^i$.  

Let $\bm{\uptheta}_S$ and  $\bm{\uptheta}_{S^i}$ represent the final GCNN filter weights learned on training set $S$ and $S^i$  respectively. Define $\Delta \bm{\uptheta} = \bm{\uptheta}_S  - \bm{\uptheta}_{S^i}  $. Using the facts that   the loss  are Lipschitz continuous and also $|\mathbf{E}[x]| \leq \mathbf{E}[|x|]$, we have,

\begin{equation*} 	
\begin{split}
& |\mathbf{E}_{\textsc{sgd}}[\ell(A_S,y) - \ell(A_{S^{i}} ,y)]|  \leq \alpha_{\ell} \mathbf{E}_{\textsc{sgd}}[| f(\mathbf{x},\bm{\uptheta}_S) -  f(\mathbf{x},\bm{\uptheta}_{S^{i}})|] \\
&  \leq \alpha_{\ell} \mathbf{E}_{\textsc{sgd}}\Big[\Big|  \sigma\Big(\sum_{\mathclap{\substack{j\in \\ \Na(\mathbf{x})}}} e_{\cdot j} \mathbf{x}_{j} \bm{\uptheta}_{S}  \Big) - \sigma\Big( \sum_{\mathclap{\substack{j \in \\ \Na(\mathbf{x})}}} \  e_{\cdot j}   \mathbf{x}_{j} \bm{\uptheta}_{S^{i}} \Big) \Big|\Big]   \\
\end{split}
\end{equation*}
\begin{equation}~\label{eq:proof_part1} 	
\begin{split}
& \text{Since activation function is also $\sigma-$Lipschitz  continuous,} \\
&   \leq \alpha_{\ell} \mathbf{E}_{\textsc{sgd}}\Big[\Big|   \sum_{\mathclap{\substack{j\in \\ \Na(\mathbf{x})}}} e_{\cdot j} \mathbf{x}_{j} \bm{\uptheta}_{S}    -   \sum_{\mathclap{\substack{j \in \\ \Na(\mathbf{x})}}} \  e_{\cdot j}   \mathbf{x}_{j} \bm{\uptheta}_{S^{i}}   \Big|\Big]   \\
& \leq \alpha_{\ell} \mathbf{E}_{\textsc{sgd}}\Big[\Big|   \sum_{\mathclap{\substack{j\in \\ \Na(\mathbf{x})}}} e_{\cdot j} \mathbf{x}_{j} (\bm{\uptheta}_{S}    -  \bm{\uptheta}_{S^{i}} )  \Big|\Big]   \\
& \leq \alpha_{\ell} \mathbf{E}_{\textsc{sgd}}\Big[ \Big|\sum_{\mathclap{\substack{j \in \\ \Na(\mathbf{x})}}} \big( e_{\cdot j }  \mathbf{x}_{j}  \big)\Big| \big(\big|\bm{\uptheta}_{S}  - \bm{\uptheta}_{S^{i}} \big|\big) \Big]   \\
& \leq \alpha_{\ell} \big| \sum_{\mathclap{\substack{j \in \\ \Na(\mathbf{x})}}} \big( e_{\cdot j }  \mathbf{x}_{j}  \big)\big| \big( \mathbf{E}_{\textsc{sgd}}\big[\big|\Delta \bm{\uptheta} \big|\big]  \big)      \\
& \leq \alpha_{\ell}  \mathbf{g}_{\lambda}   \mathbf{E}_{\textsc{sgd}}\big[\big|\Delta \bm{\uptheta} \big|\big]     \\
\end{split}
\end{equation}

where 
$\mathbf{g}_\lambda$ is defined as $  \mathbf{g}_\lambda:=\underset{\mathbf{x}}{\sup}  \hspace{0.5em} \Big|\sum_{j \in  \Na(\mathbf{x})}   e_{\cdot j }  \mathbf{x}_{j}  \Big| $.
We will bound $\mathbf{g}_\lambda$ in terms of the largest absolute eigenvalue of the graph convolution filter $g(\mathbf{L})$ later. Note that $\sum_{j \in  \Na(\mathbf{x})}   e_{\cdot j } \mathbf{x}_{j}$  is nothing but a graph convolution operation. As such, reducing $\mathbf{g}_\lambda$   will be the contributing factor in improving the generalization performance. \\


\noindent \textbf{Proof Part II (SGD  Based Bounds For GCNN Weights)}: 
What remains is to bound $\mathbf{E}_{\textsc{sgd}}[|\Delta \bm{\uptheta}|]$ due to the randomness inherent in SGD. This is proved through a series of three lemmas. We first note that on a given training set $S$, a GCNN   minimizes  the following objective function,
\begin{equation}~\label{eq:obj}
\begin{aligned}
& \underset{\bm{\uptheta}}{\min} \hspace{1em}\La \big(f(\mathbf{x},\bm{\uptheta}_S),y\big) =\frac{1}{m}\sum\limits_{i=1}^{m}  \ell\big( f(\mathbf{x},\bm{\uptheta}_S),y_i\big) \\
\end{aligned}
\end{equation}
For this, at each iteration $t$, SGD performs the following update: 
\begin{equation}
\begin{aligned}
\bm{\uptheta}_{S,t+1}=\bm{\uptheta}_{S,t}-\eta  \nabla \ell\big( f(\mathbf{x}_{i_t},\bm{\uptheta}_{S,t}),y_{i_t}\big) \label{eq:SGD-update}
\end{aligned}
\end{equation}
where  $\eta >0$ is the learning rate. 

Given two sequences of the weight parameters, $\{\bm{\uptheta}_{S,0}$ ,  $\bm{\uptheta}_{S,1}$  $,\dots,$ $\bm{\uptheta}_{S,T}\}$ and  $\{\bm{\uptheta}_{S^i,0},$  $\bm{\uptheta}_{S^i,1}$ $,\dots,$  $\bm{\uptheta}_{S^i,T} \}$, learned by the GCCN running SGD on $S$ and $S^{i}$, respectively, we first find a bound on
$\Delta\bm{\uptheta}_{t}:=|\bm{\uptheta}_{S,t} -\bm{\uptheta}_{S^i,t}|$ at each iteration step $t$ of SGD.   

There are two scenarios to consider  1) At step $t$, SGD  picks a sample $\mathbf{z}_t=(\mathbf{x},y)$ which is identical in $Z$ and $Z'$, and   occurs with probability $(m-1)/m$. From Equation~(\ref{eq:SGD-update}), we have  $|\Delta\bm{\uptheta}_{t+1}| \leq |  \Delta\bm{\uptheta}_{t}|+\eta |\nabla\ell\big( f(\mathbf{x},\bm{\uptheta}_{S,t}),y\big) - \ell\big( f(\mathbf{x},\bm{\uptheta}_{S,t}),y\big)|$. We bound this term in Lemma~\ref{lemma:sgd_term1} below  2) At step $t$, SGD  picks the only samples  
that  $Z$ and $Z'$ differ,   $\mathbf{z}_t=(\mathbf{x}_i,y_i)$ and $\mathbf{z}'_t=(\mathbf{x}'_i,y'_i)$ which occurs with probability $1/m$. Then $|\Delta\bm{\uptheta}_{t+1}| \leq | \Delta\bm{\uptheta}_{t} | + \eta |\nabla\ell\big( f(\mathbf{x}_i,\bm{\uptheta}_{S,t}),y_i\big) - \ell\big( f(\mathbf{x}'_i,\bm{\uptheta}_{S,t}),y'_i\big)|$. We bound the second term in Lemma~\ref{lemma:sgd_term2} below.

\begin{lemma}[\textbf{GCNN Same Sample Loss Stability Bound}]\label{lemma:sgd_term1}  \textit{The   loss-derivative bound difference   of (single-layer) GCNN models trained with SGD algorithm for $T$ iterations on two training datasets $S$ and $S^i$ respectively, with respect to the same sample     is given by,}
	$$ \Big|   \nabla \ell\big(f(\mathbf{x},\bm{\uptheta}_{S, t} ),y \big)   -      \nabla \ell\big(f(\mathbf{x},\bm{\uptheta}_{S^{i}, t} ),y \big)   \Big| \leq \nu_{\ell}  \nu_{\sigma} \mathbf{g}_{\lambda}^2  |   \Delta\bm{\uptheta}_{t}|. $$
\end{lemma}
\noindent \textbf{Proof}: The first order derivative of a single-layer the GCNN output function, $f(\mathbf{x},\bm{\uptheta} ) = \sigma ( \sum_{j \in \\ \Na}  e_{\cdot j}  \mathbf{x}_{j}  \bm{\uptheta} )$,  is given by,

\begin{equation}~\label{eq:gcnn_derivative}	
\frac{\partial f(\mathbf{x},\bm{\uptheta} )}{\partial \bm{\uptheta} }  = \sigma' \Big( \sum_{\mathclap{\substack{j\in \\ \Na(\mathbf{x})}}}  e_{\cdot j}  \mathbf{x}_{j}  \bm{\uptheta} \Big)  \sum_{\mathclap{\substack{j\in \\ \Na(\mathbf{x}) }}}  e_{\cdot j}  \mathbf{x}_{j},   
\end{equation}
where $\nabla \sigma(\cdot)$ is the first order derivative of the activation function.

Using Equation (\ref{eq:gcnn_derivative}) and the fact that the loss function is Lipschitz continuous and smooth, we have,

\begin{equation} 	
\begin{split}
& \Big|   \nabla \ell\big(f(\mathbf{x},\bm{\uptheta}_{S, t} ),y \big)   -      \nabla \ell\big(f(\mathbf{x},\bm{\uptheta}_{S^{i}, t} ),y \big)   \Big| \nonumber \leq \\ & \hspace{10em} \nu_{\ell} \big| \nabla f(\mathbf{x},\bm{\uptheta}_{S, t} ) - \nabla f(\mathbf{x},\bm{\uptheta}_{S^{i},t } ) \big|  \\
& \leq \nu_{\ell} \Big| \nabla  \sigma\Big( \sum_{\mathclap{\substack{j\in \\ \Na(\mathbf{x})}}}  e_{\cdot j}  \mathbf{x}_{j}  \bm{\uptheta}_{S, t} \Big)  \sum_{\mathclap{\substack{j\in \\ \Na(\mathbf{x}) }}}  e_{\cdot j}  \mathbf{x}_{j} - \nonumber\\ 
& \hspace{10em} \nabla \sigma\Big( \sum_{\mathclap{\substack{j\in \\ \Na(\mathbf{x})}}}  e_{\cdot j}  \mathbf{x}_{j}  \bm{\uptheta}_{S^{i}, t} \Big)  \sum_{\mathclap{\substack{j\in \\ \Na(\mathbf{x}) }}}  e_{\cdot j}  \mathbf{x}_{j} \Big|  \\
& \leq \nu_{\ell}\Big( \big| \sum_{\mathclap{\substack{j\in \\ \Na(\mathbf{x}) }}}  e_{\cdot j}  \mathbf{x}_{j} \big|  \Big) \Big| \nabla \sigma\Big( \sum_{\mathclap{\substack{j\in \\ \Na(\mathbf{x})}}}  e_{\cdot j}  \mathbf{x}_{j}  \bm{\uptheta}_{S, t} \Big)   - \nabla  \sigma\Big( \sum_{\mathclap{\substack{j\in \\ \Na(\mathbf{x})}}}  e_{\cdot j}  \mathbf{x}_{j}  \bm{\uptheta}_{S^{i}, t} \Big)    \Big|  \\
& \text{Since the activation function is    Lipschitz continuous and smooth,  } \\
& \text{and plugging  $\big| \sum_{\mathclap{\substack{j\in \\ \Na(\mathbf{x}) }}}  e_{\cdot j}  \mathbf{x}_{j} \big| \leq \mathbf{g}_\lambda$, we get,  } \\
& \leq \nu_{\ell}  \nu_{\sigma} \mathbf{g}_{\lambda} \Big|  \Big( \sum_{\mathclap{\substack{j\in \\ \Na(\mathbf{x})}}}  e_{\cdot j}  \mathbf{x}_{j}  \bm{\uptheta}_{S, t} \Big)   -  \Big( \sum_{\mathclap{\substack{j\in \\ \Na(\mathbf{x})}}}  e_{\cdot j}  \mathbf{x}_{j}  \bm{\uptheta}_{S^{i}, t} \Big)    \Big|  \\
& \leq \nu_{\ell}  \nu_{\sigma} \mathbf{g}_{\lambda} \Big( \big| \sum_{\mathclap{\substack{j\in \\ \Na(\mathbf{x})}}}  e_{\cdot j}  \mathbf{x}_{j}  \big| \Big)   |\bm{\uptheta}_{S, t} - \bm{\uptheta}_{S^{i}, t}|  \\
& \leq \nu_{\ell}  \nu_{\sigma} \mathbf{g}_{\lambda}^2 | \Delta\bm{\uptheta}_{t}|  
\end{split}
\end{equation}
This completes the proof of Lemma~\ref{lemma:sgd_term1}. \\

\noindent \textbf{Note}: Without the $\sigma-$smooth assumption, it would not be possible to  derive the above bound in terms of $|\Delta\bm{\uptheta}_{t}|$ which is  necessary for showing the uniform stability. Unfortunately, this constraint excludes RELU activation from our analysis.

\begin{lemma}[\textbf{GCNN Different Sample Loss Stability Bound}]\label{lemma:sgd_term2}
	\textit{The   loss-derivative bound difference   of (single-layer) GCNN models trained with SGD algorithm for $T$ iterations on two training datasets $S$ and $S^i$ respectively,  with respect to  the  different samples     is given by,}
	$$ \Big|   \nabla \ell\big(f(\mathbf{x}_i,\bm{\uptheta}_{S, t} ),y_i \big)   -      \nabla \ell\big(f(\mathbf{x}'_i,\bm{\uptheta}_{S^{i}, t} ),y'_i \big)   \Big| \leq 2\nu_{\ell} \alpha_{\sigma} \mathbf{g}_{\lambda}. $$
\end{lemma}
\noindent \textbf{Proof}: 
Again using Equation~(\ref{eq:gcnn_derivative})  and the fact that the loss \& activation function is Lipschitz continuous and smooth, and for any $a$, $b$, $|a-b| \leq |a|+|b|$, we have, 

\begin{equation} 	
\begin{split}
& \Big|   \nabla \ell\big(f(\mathbf{x},\bm{\uptheta}_{S, t}),y \big)   -      \nabla \ell\big(f(\mathbf{x}^{'},\bm{\uptheta}_{S^{i}, t}),y^{'} \big)   \Big|     \leq   \\ & \hspace{10em} \nu_{\ell} \big| \nabla f(\mathbf{x},\bm{\uptheta}_{S, t}) - \nabla f(\mathbf{x}^{'},\bm{\uptheta}_{S^{i}, t})\big| \\
& \leq \nu_{\ell} \Big| \nabla \sigma \Big( \sum_{\mathclap{\substack{j\in \\ \Na(\mathbf{x})}}}  e_{\cdot j}  \mathbf{x}_{j}  \bm{\uptheta}_{S, t} \Big)  \sum_{\mathclap{\substack{j\in \\ \Na(\mathbf{x}) }}}  e_{\cdot j}  \mathbf{x}_{j} - \nabla \sigma \Big( \sum_{\mathclap{\substack{j\in \\ \Na(\mathbf{x}^{'})}}}  e_{\cdot j}  \mathbf{x}_{j}^{'}  \bm{\uptheta}_{S^{i}, t} \Big)  \sum_{\mathclap{\substack{j\in \\ \Na(\mathbf{x}^{'}) }}}  e_{\cdot j}  \mathbf{x}_{j}^{'} \Big|   \\
& \leq \nu_{\ell}\Big| \nabla\sigma \Big( \sum_{\mathclap{\substack{j\in \\ \Na(\mathbf{x})}}}  e_{\cdot j}  \mathbf{x}_{j}  \bm{\uptheta}_{S, t} \Big)  \sum_{\mathclap{\substack{j\in \\ \Na(\mathbf{x}) }}}  e_{\cdot j}  \mathbf{x}_{j} \Big|  + \\ & \hspace{1.5em} \nu_{\ell}\Big| \nabla \sigma\Big( \sum_{\mathclap{\substack{j\in \\ \Na(\mathbf{x}^{'})}}}  e_{\cdot j}  \mathbf{x}_{j}^{'}  \bm{\uptheta}_{S^{i}, t} \Big)  \sum_{\mathclap{\substack{j\in \\ \Na(\mathbf{x}^{'}) }}}  e_{\cdot j}  \mathbf{x}_{j}^{'} \Big|  \\
& \text{Using the fact that  the first order derivative is bounded,  } \\
& \leq 2\nu_{\ell} \alpha_{\sigma} \mathbf{g}_{\lambda}  \\
\end{split}
\raisetag{6\normalbaselineskip}
\end{equation}
This completes the proof of Lemma~\ref{lemma:sgd_term2}. \\


Summing over all iteration steps, and taking expectations over all possible sample sequences $Z$, $Z'$ from $S$ and $S^i$, we have 

\begin{lemma}[\textbf{GCNN SGD Stability Bound}]\label{lemma:sgd_terms} \textit{ Let the loss \& activation functions be  Lipschitz-continuous and smooth. Let $\bm{\uptheta}_{S,T}$ and $\bm{\uptheta}_{S^i,T}$ denote the graph filter parameters of (single-layer) GCNN models	 trained using SGD for $T$ iterations on two training datasets  $S$ and $S^{i}$, respectively. Then the expected difference in   the filter  parameters  is bounded by,}
	$$ \mathbf{E}_{\textsc{sgd}}  \big[\big|\Delta \bm{\uptheta}_{S,T} -\bm{\uptheta}_{S^i,T}  | \big]   \leq    \frac{2\eta \nu_{\ell} \alpha_{\sigma}  \mathbf{g}_{\lambda}  }{m}  \sum_{t=1}^{T} \big( 1 + \eta\nu_{\ell}  \nu_{\sigma} \mathbf{g}_\lambda^2 \big)^{t-1}  $$
\end{lemma}
\noindent \textbf{Proof}: 
From Equation~(\ref{eq:SGD-update}) and  taking into account the probabilities of the two scenarios considered in Lemma~\ref{lemma:sgd_term1} and Lemma~\ref{lemma:sgd_term2} at step $t$, we have,
\begin{equation*}
\begin{split}
& \mathbf{E}_{\textsc{sgd}}  \big[\big|\Delta \bm{\uptheta}_{t+1} | \big]   \leq \Big(1-\frac{1}{m}\Big) \mathbf{E}_{\textsc{sgd}}  \Big[ \Big| \Big(\bm{\uptheta}_{S, t}  - \eta  \nabla \ell\big(f(\mathbf{x},\bm{\uptheta}_{S, t}),y \big) \Big) - \\ & \hspace{1.5em} \Big(\bm{\uptheta}_{S^i, t}   - \eta  \nabla \ell\big(f(\mathbf{x},\bm{\uptheta}_{S^{i}, t}), y \big) \Big) \Big| \Big] + \Big(\frac{1}{m}  \Big) \mathbf{E}_{\textsc{sgd}}  \Big[ \Big| \Big(\bm{\uptheta}_{S, t}    - \\ & \hspace{3em}  \eta  \nabla \ell\big(f(\mathbf{x}^{'},\bm{\uptheta}_{S,t}),y^{'} \big) \Big)  -   \Big(\bm{\uptheta}_{S^i, t}   - \eta  \nabla \ell\big(f(\mathbf{x}^{''},\bm{\uptheta}_{S^{i}, t}),y^{''} \big) \Big) \Big| \Big]  \\
\end{split}
\end{equation*}
\begin{equation}~\label{eq:sgd_terms} 		
\begin{split}
& \leq \Big(1-\frac{1}{m}\Big) \mathbf{E}_{\textsc{sgd}}  \big[|\Delta \bm{\uptheta}_{t} | \big] + \Big(1-\frac{1}{m}\Big) \eta\mathbf{E}_{\textsc{sgd}}  \Big[ \Big|   \nabla \ell\big(f(\mathbf{x},\bm{\uptheta}_{S, t}),y \big)   - \\  &  \hspace{1.5em}  \nabla \ell\big(f(\mathbf{x},\bm{\uptheta}_{S^{i}, t}),y\big)  \Big| \Big]  +  \Big(\frac{1}{m}\Big) \mathbf{E}_{\textsc{sgd}}  \big[|\Delta \bm{\uptheta}_{t} | \big]  + \\ & \hspace{1.5em} \Big(\frac{1}{m}\Big) \eta\mathbf{E}_{\textsc{sgd}}  \Big[ \Big|   \nabla \ell\big(f(\mathbf{x}^{'},\bm{\uptheta}_{S, t}),y^{'} \big)   -  \nabla \ell\big(f(\mathbf{x}^{''},\bm{\uptheta}_{S^{i}, t}),y^{''} \big)  \Big| \Big]  \\ 
& = \mathbf{E}_{\textsc{sgd}}  \big[|\Delta \bm{\uptheta}_{t} | \big] + \\ & \hspace{1.5em} \Big(1-\frac{1}{m}\Big) \eta\mathbf{E}_{\textsc{sgd}}  \Big[ \Big|   \nabla \ell\big(f(\mathbf{x},\bm{\uptheta}_{S, t}),y \big)   -      \nabla \ell\big(f(\mathbf{x},\bm{\uptheta}_{S^{i}, t}),y\big)   \Big| \Big]   + \\ &    \hspace{1.5em}  \Big(\frac{1}{m}\Big) \eta\mathbf{E}_{\textsc{sgd}}  \Big[ \Big|  \big(\nabla \ell\big(f(\mathbf{x}^{'},\bm{\uptheta}_{S, t}),y^{'} \big) \Big) -  \big(\nabla \ell\big(f(\mathbf{x}^{''},\bm{\uptheta}_{S^{i}, t}),y^{''} \big) \Big) \Big| \Big].  \\ 
\end{split}
\raisetag{8\normalbaselineskip}
\end{equation} \\

Plugging the bounds in	Lemma~\ref{lemma:sgd_term1} and Lemma~\ref{lemma:sgd_term2} into Equation~(\ref{eq:sgd_terms}),  we have,
\begin{equation*} 	
\begin{split}
& \mathbf{E}_{\textsc{sgd}}  \big[\big|\Delta \bm{\uptheta}_{t+1} | \big] \leq \mathbf{E}_{\textsc{sgd}}  \big[|\Delta \bm{\uptheta}_{t} | \big] + \Big(1-\frac{1}{m}\Big) \eta\nu_{\ell}  \nu_{\sigma}   \mathbf{g}_{\lambda}^2    \mathbf{E}_{\textsc{sgd}}   [ |     \bm{\uptheta}_{t}|  ]  \\ & \hspace{10em}  + \Big(\frac{1}{m}\Big) 2\eta \nu_{\ell} \alpha_{\sigma} \mathbf{g}_{\lambda} \\ 
& = \Big( 1 + \big(1-\frac{1}{m}\big) \eta\nu_{\ell}  \nu_{\sigma} \mathbf{g}_{\lambda}^2 \Big)   \mathbf{E}_{\textsc{sgd}}   [ |     \bm{\uptheta}_{t}|  ]    + \frac{2\eta \nu_{\ell} \alpha_{\sigma} \mathbf{g}_{\lambda} }{m}  \\ 
& \leq \Big( 1 + \eta\nu_{\ell}  \nu_{\sigma} \mathbf{g}_{\lambda}^2  \Big)   \mathbf{E}_{\textsc{sgd}}   [ |     \bm{\uptheta}_{t}|  ]    + \frac{2\eta \nu_{\ell} \alpha_{\sigma} \mathbf{g}_{\lambda}   }{m}.  \\
\end{split}
\end{equation*}

Lastly,   solving the $\mathbf{E}_{\textsc{sgd}} \big[\big|\Delta\bm{\uptheta}_{t} | \big]$ first order recursion yields, 
\begin{equation*} 	
\begin{split}	
& \mathbf{E}_{\textsc{sgd}}  \big[\big|\Delta \bm{\uptheta}_{T} | \big]   \leq    \frac{2\eta \nu_{\ell} \alpha_{\sigma} \mathbf{g}_{\lambda} }{m}  \sum_{t=1}^{T} \big( 1 + \eta\nu_{\ell}  \nu_{\sigma} \mathbf{g}_\lambda^2 \big)^{t-1}  \\
\end{split}
\end{equation*}
This completes the proof of Lemma~\ref{lemma:sgd_terms}. \\

\noindent \textbf{Bound on $\mathbf{g}_{\lambda}$: } We now bound $\mathbf{g}_{\lambda}$ in terms of the largest absolute eigenvalue of the graph filter matrix $g(\mathbf{L})$. We first note that at each node $\mathbf{x}$, the ego-graph $G_\mathbf{x}$ ego-graph can be represented as a sub-matrix of $g(\mathbf{L})$. 
Let $g_{\mathbf{x}}(\mathbf{L}) \in \mathbb{R}^{q \times q}$ be the submatrix of  $g(\mathbf{L})$ whose row and column indices are from the set $\{j \in  \Na(\mathbf{x})\}$. The ego-graph size is   $q=|\Na(\mathbf{x})|$. We use $\mathbf{h}_{\mathbf{x}} \in \mathbb{R}^{q}$ to denote the graph signals (node features) on the ego-graph $G_\mathbf{x}$.  Without loss of generality, we will assume that node $\mathbf{x}$ is represented by index $0$ in   $G_\mathbf{x}$. Thus, we can compute $ \sum_{j \in \Na(\mathbf{x})} e_{\cdot j}  \mathbf{x}_{j} = [g_{\mathbf{x}}(\mathbf{L}) \mathbf{h}_{\mathbf{x}}]_0$, a scalar value. Here $[\cdot]_0 \in \mathbb{R}$ represents the value of a vector at index 0, i.e., corresponding to  node $\mathbf{x}$. Then the following holds (assuming the graph signals are normalized, i.e., $\| \mathbf{h}_{\mathbf{x}} \|_{2}=1$),
\begin{equation}~\label{eq:g_lambda} 	
\begin{split}
|[g_{\mathbf{x}}(\mathbf{L}) \mathbf{h}_{\mathbf{x}}]_0| \leq \|g_{\mathbf{x}}(\mathbf{L})  \mathbf{h}_{\mathbf{x}} \|_{1} \leq \|\|g_{\mathbf{x}}(\mathbf{L})\|_2  \|\mathbf{h}_{\mathbf{x}} \|_{2} =   \lambda_{G_\mathbf{x}}^{\max}   
\end{split} 
\end{equation}
where the second inequality follows from Cauchy–Schwarz Inequality, and $\|M\|_2=\sup_{\|x\|_2=1} \|Mx\|_2= \sigma_{max}(M)$ is the matrix operator norm  and $\sigma_{max}(M)$ is the largest singular value of matrix $M$. For a normal matrix $M$ (such as a symmetric graph filter $g(L)$),  $\sigma_{max}(M)=\max |\lambda(M)|$, the largest absolute eigenvalue of $M$. 

\begin{lemma}[\textbf{Ego-Graph Eigenvalue Bound}]\label{lemma:eigenvalue} \textit{Let $G=(V, E)$ be a (un)directed graph with (either symmetric or non-negative) weighted adjacency matrix $g(\mathbf{L})$  and $\lambda_{G}^{\max}$ be the maximum absolute eigenvalue of $g(\mathbf{L})$. Let $G_\mathbf{x}$ be the ego-graph of a node $\mathbf{x} \in V$   with corresponding maximum absolute eigenvalue $ \lambda_{G_\mathbf{x}}^{\max}  $. Then the following eigenvalue (singular value) bound holds  $\forall \mathbf{x}$,  }
	$$  \lambda_{G_\mathbf{x}}^{\max} \leq \lambda_{G}^{\max}    $$
\end{lemma}
\noindent \textbf{Proof}:  Notice that $g_{\mathbf{x}}(\mathbf{L})$ is   the adjacency matrix of $G_\mathbf{x}$ which also happens to be  the principal submatrix of   $g(\mathbf{L})$. As a result, above bound holds from the eigenvalue interlacing theorem for normal/Hermitian matrices and their principal submatrices~\cite{laffey2008spectra, haemers1995interlacing}. 


Finally, plugging $\mathbf{g}_\lambda \leq \lambda_{G}^{\max}$ and Lemma~\ref{lemma:sgd_terms} into Equation~(\ref{eq:proof_part1}) yields the following remaining result,

\begin{equation*} 	
\begin{split}
& 2\beta_{m}  \leq  \alpha_{\ell}  \lambda_{G}^{\max}   \mathbf{E}_{\textsc{sgd}}\big[\big|\Delta \bm{\uptheta} \big|\big]        \\
& \beta_{m}  \leq   \frac{ \eta \alpha_{\ell}   \alpha_{\sigma} \nu_{\ell}  (\lambda_{G}^{\max})^2 \sum_{t=1}^{T} \big( 1 + \eta\nu_{\ell}  \nu_{\sigma} (\lambda_{G}^{\max})^2 \big)^{t-1}}{m}    \\  
&   \beta_{m}  \leq  \frac{1}{m}\BigO \Big( (\lambda_{G}^{\max})^{2T}\Big)  \hspace{2em} \forall T \geq 1  \\ 
\end{split} 
\end{equation*}

This completes the full proof of Theorem~\ref{thm:gcnn-uniform-stability}.

%
%


\section{Revisiting  Graph Convolutional Neural Network     Architecture}\label{sec:implication}

\begin{figure*}[!ht]
	\centering
	\begin{subfigure}[b]{0.32\textwidth}
		\includegraphics[width=1\textwidth]{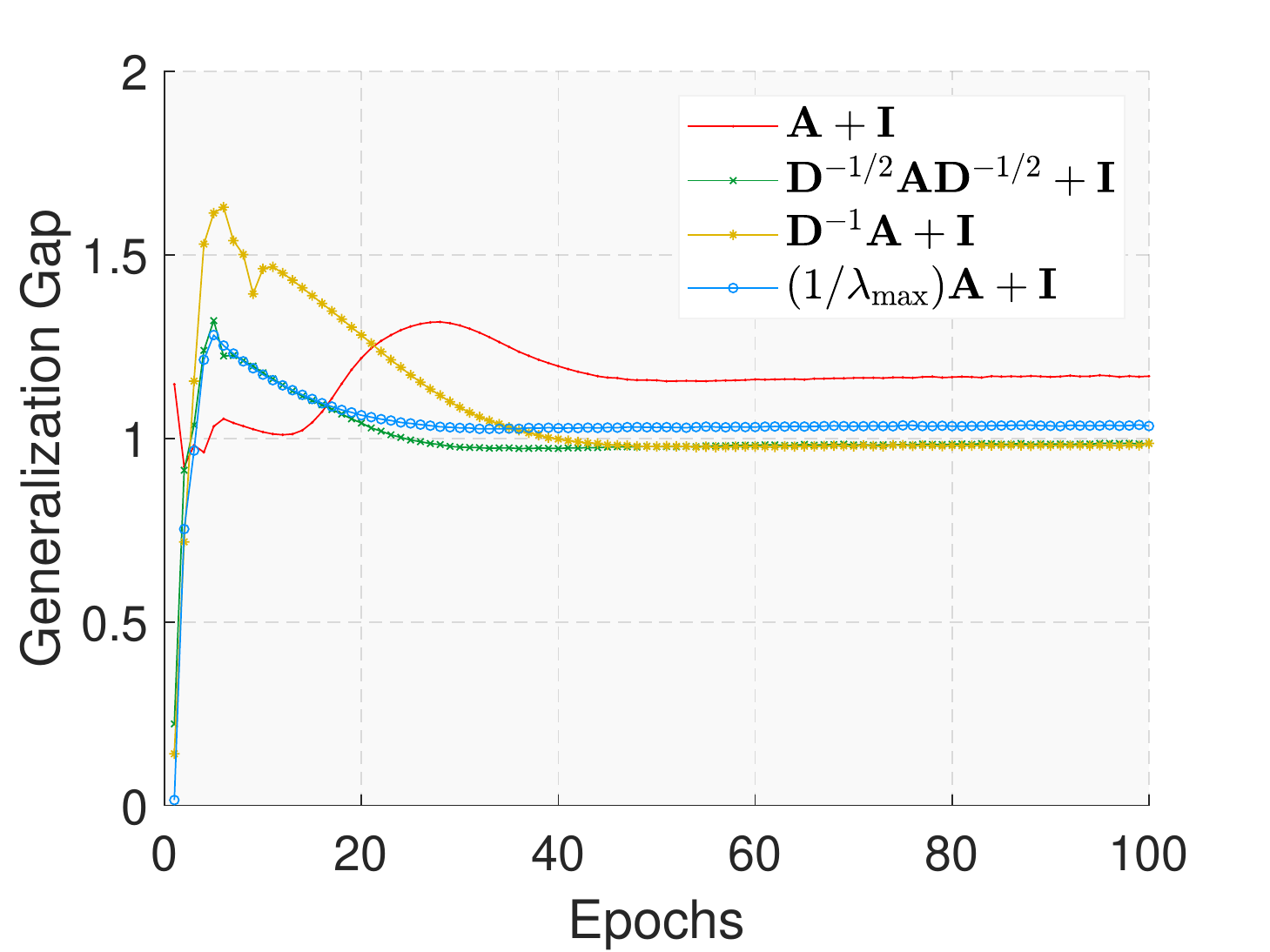}
		\caption{Generlization Gap on Citeseer Dataset}
		\label{fig:transit-reg-morning-low-dim}
	\end{subfigure}
	~~~
	\begin{subfigure}[b]{0.32\textwidth}
		\includegraphics[width=1\textwidth]{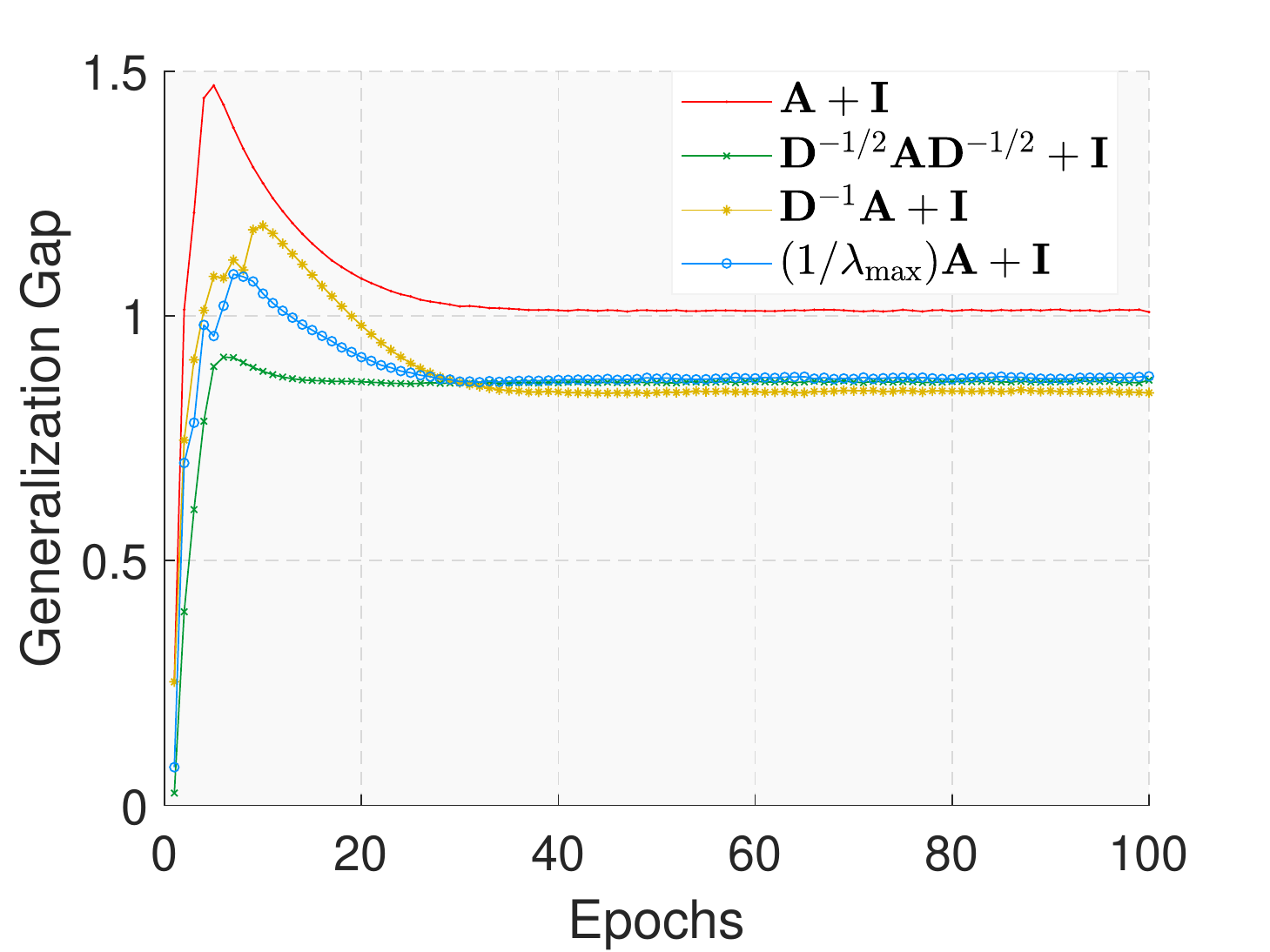}
		\caption{Generlization Gap on Cora Dataset}
		\label{fig:transit-reg-evening-low-dim}
	\end{subfigure}	
	~~~
	\begin{subfigure}[b]{0.32\textwidth}
		\includegraphics[width=1\textwidth]{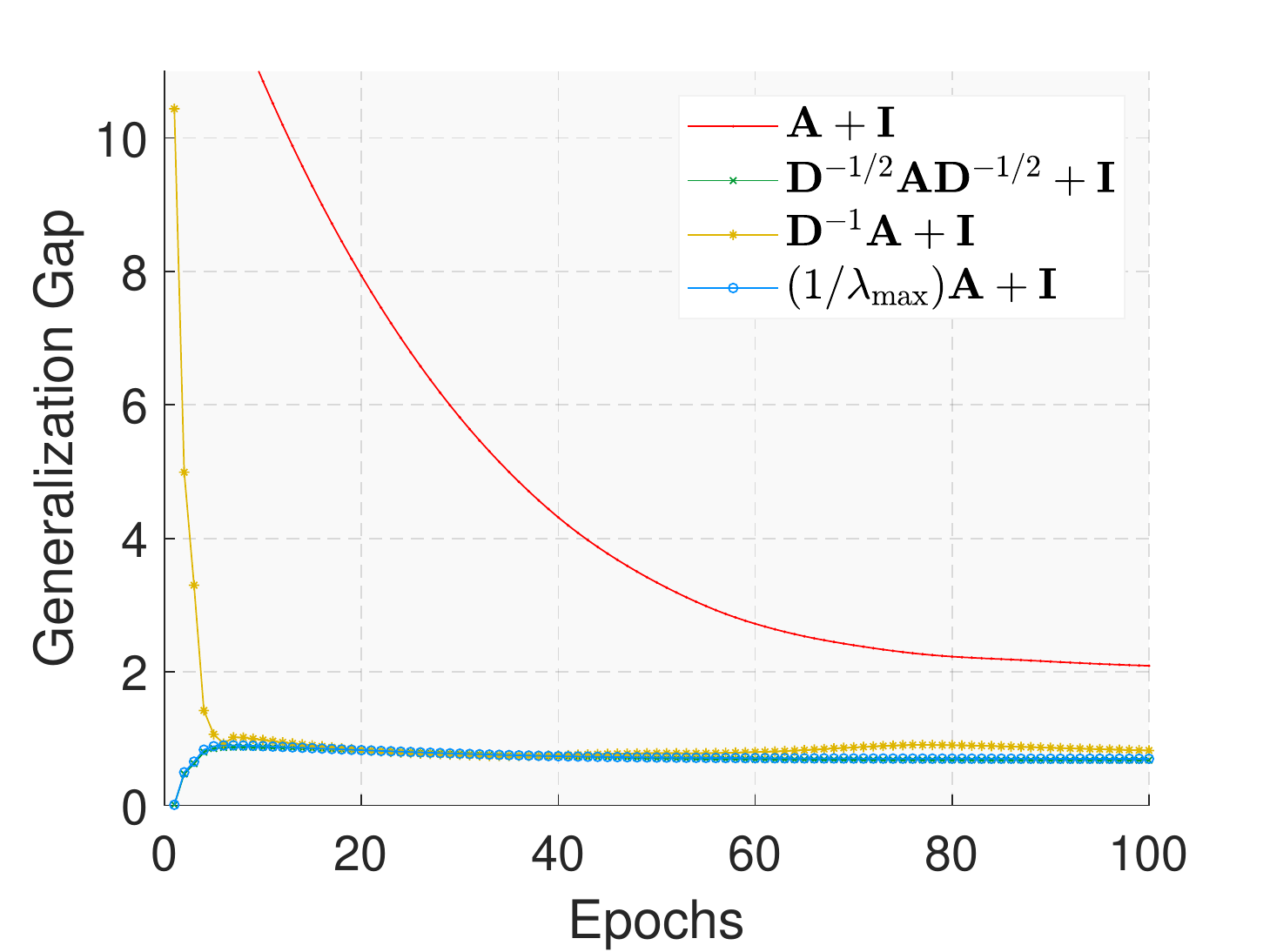}
		\caption{Generlization Gap on Pubmed Dataset}
		\label{fig:transit-reg-evening-low-dim}
	\end{subfigure}	
	\caption{The above figures show the generalziation gap for three datasets. The generlization gap is measured with respect to the loss function, i.e., \textrm{|(training error $-$ test error)|}. In this experiment,  the cross-entropy loss is used.}\label{fig:gen_gap}
\end{figure*}

\begin{figure*}[!ht]
	\centering
	\begin{subfigure}[b]{0.30\textwidth}
		\includegraphics[width=1\textwidth]{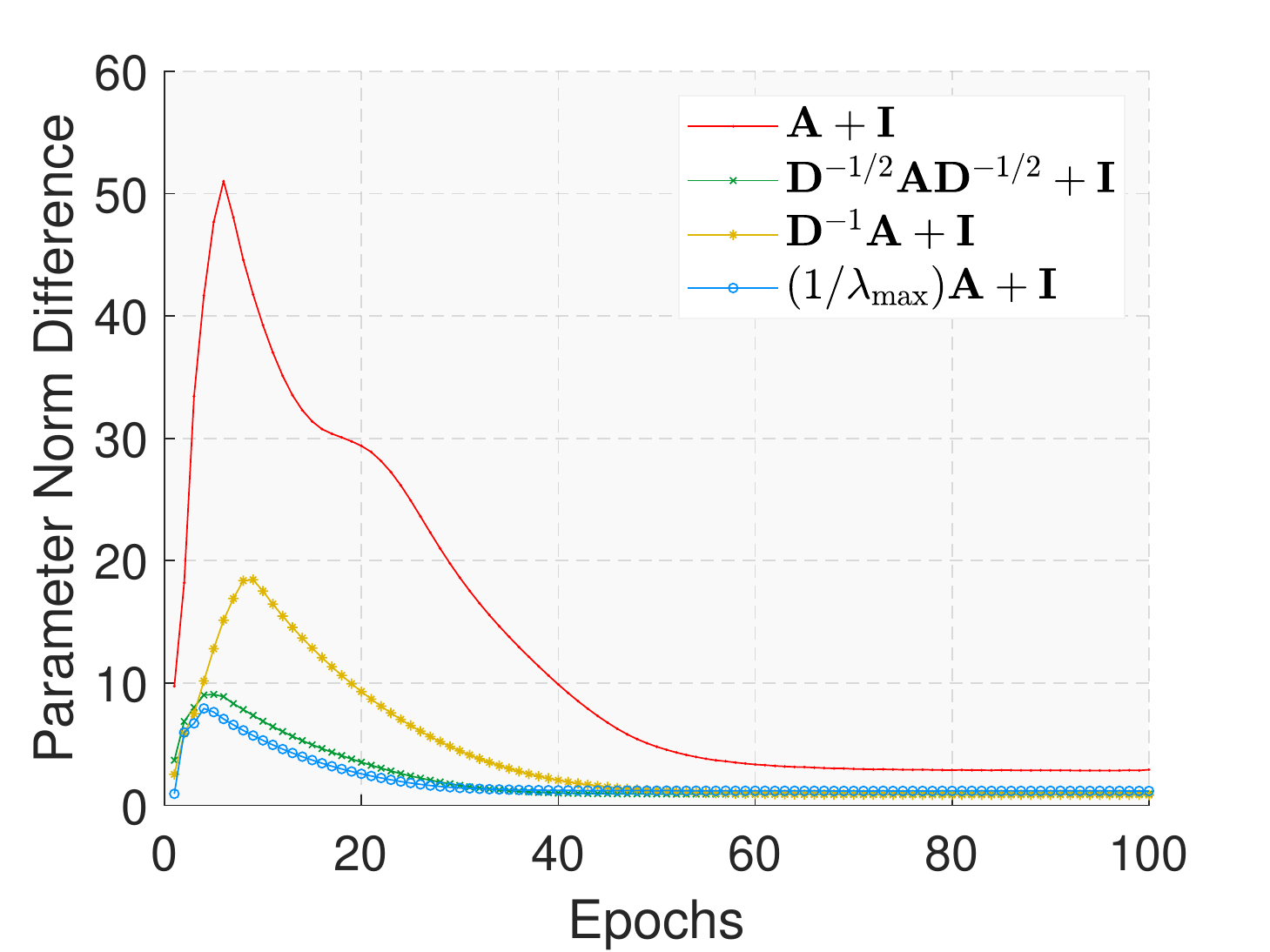}
		\caption{Parameter $L2-$Norm Diff on Citeseer  }
		\label{fig:transit-reg-morning-low-dim}
	\end{subfigure}
	~~~
	\begin{subfigure}[b]{0.30\textwidth}
		\includegraphics[width=1\textwidth]{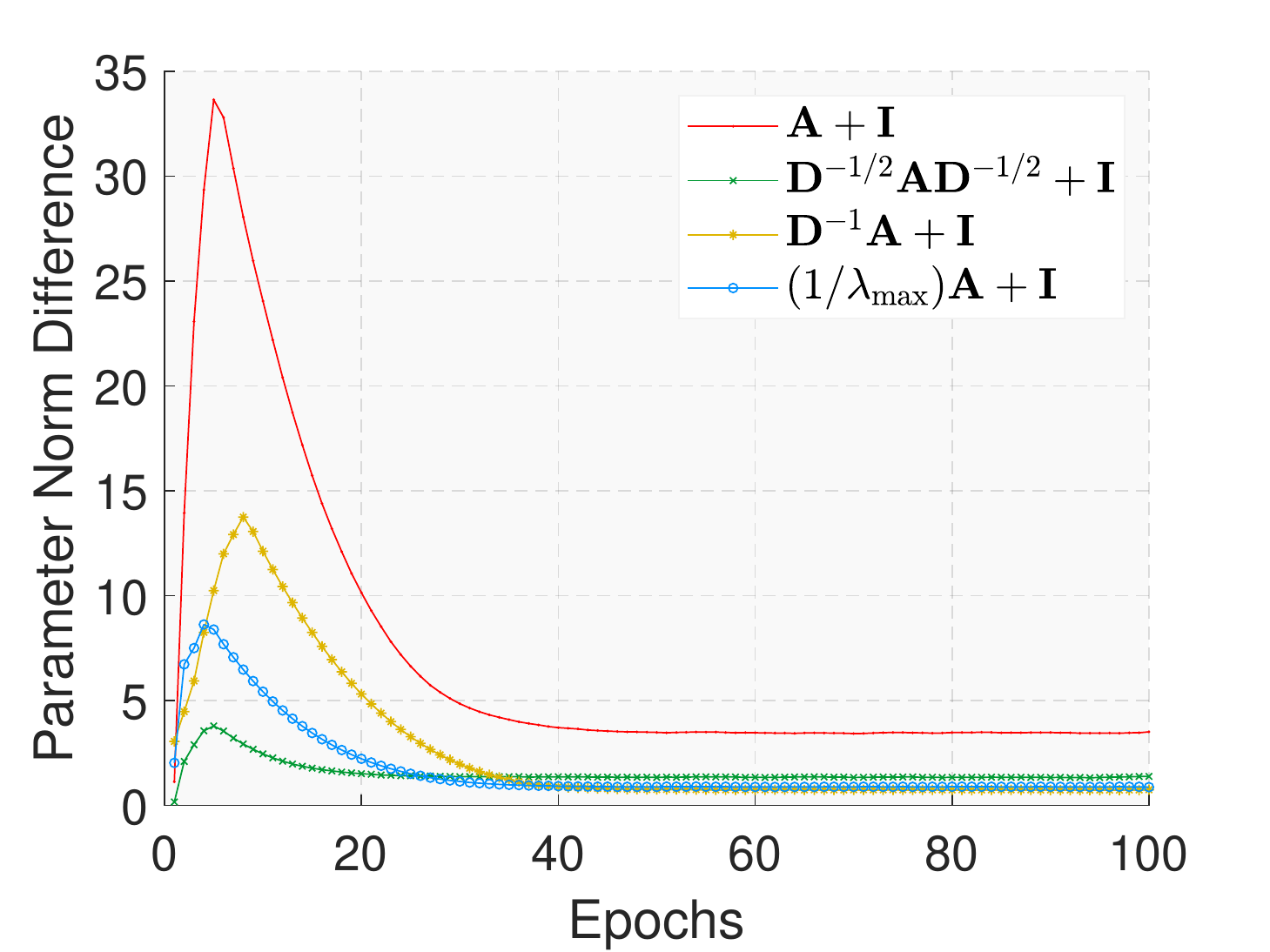}
		\caption{Parameter $L2-$Norm Diff on Cora  }
		\label{fig:transit-reg-evening-low-dim}
	\end{subfigure}	
	~~~
	\begin{subfigure}[b]{0.30\textwidth}
		\includegraphics[width=1\textwidth]{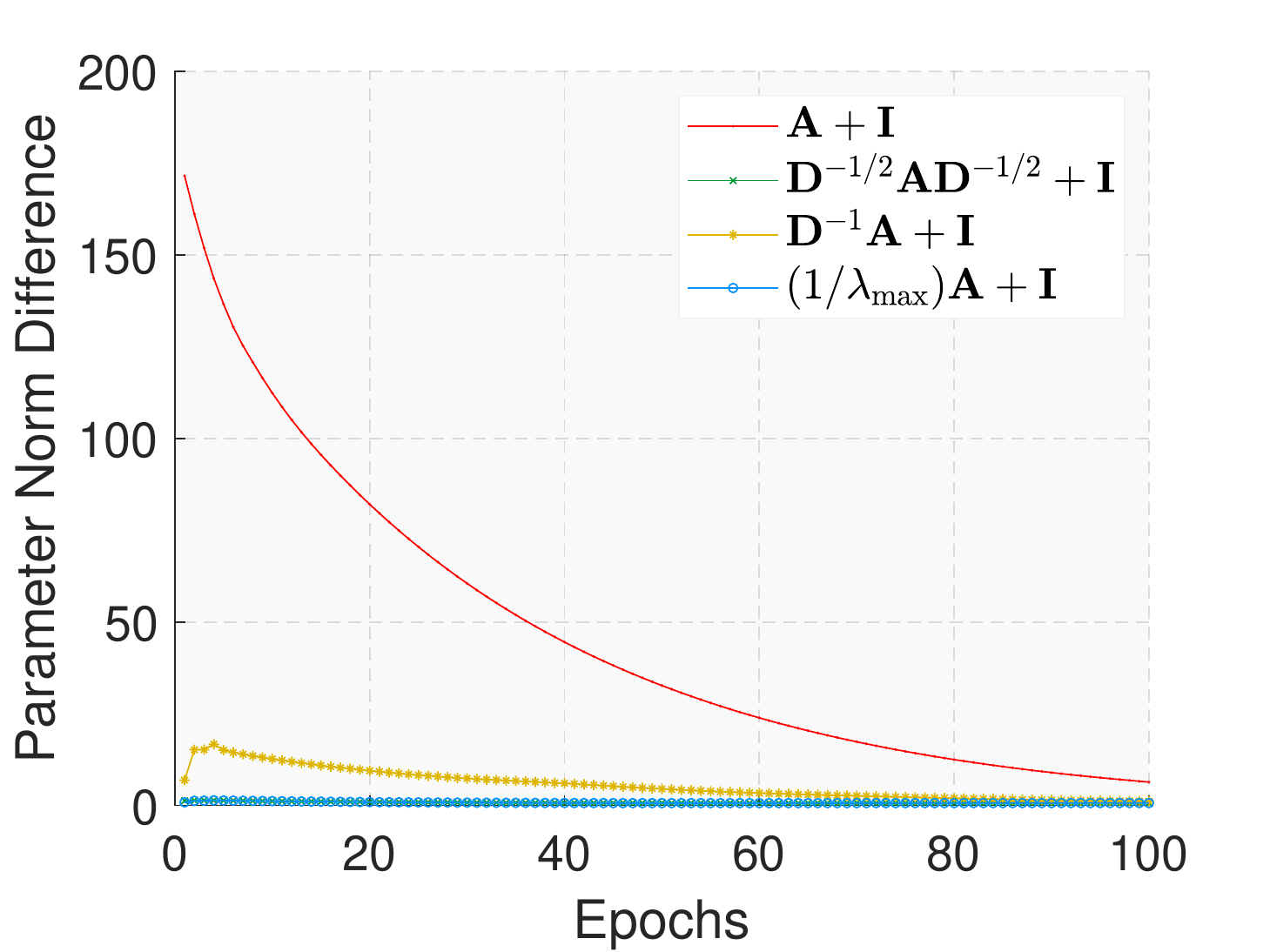}
		\caption{Parameter $L2-$Norm Diff on Pubmed  }
		\label{fig:transit-reg-evening-low-dim}
	\end{subfigure}	
	\caption{The above figures show the divergence in weight parameters of a single layer GCNN measured using $L2-$norm on the three datasets. We surgically alter one sample point  at index $i=0$ in the training set $S$ to generate $S^{i}$ and run the SGD algorithm.}\label{fig:parms_norm}
\end{figure*}

In this section, we discuss the implication of our results in designing   graph convolution filters and revisit the importance of employing batch-normalization layers in GCNN network.  \\

\noindent \textbf{Unnormalized   Graph Filters}: One of the most popular graph convolution filters is $g(\mathbf{L}) = \mathbf{A} + \mathbf{I}$~\cite{xu2018powerful}. The eigen spectrum of the unnormalized $\mathbf{A}$ is bounded by $\BigO(N)$.   This is concerning as now $\mathbf{g}_\lambda$   is  bounded by $\BigO(N)$ and as $m$ becomes close to $N$, $\beta_m$   tend towards $\BigO(N^{c})$ complexity with $c\geq 0$. As a result, the generalization gap of such a GCNN model is not   guaranteed to converge. \\

\noindent \textbf{Normalized   Graph  Filters}: Numerical instabilities with the unnormalized adjacency matrix have already been suspected in~\cite{kipf2016semi}. Therefore, the symmetric normalized graph filter has been adopted: $g(\mathbf{L}) = \mathbf{D}^{-1/2}\mathbf{A}\mathbf{D}^{-1/2} + \mathbf{I}$ . The eigen spectrum of  $\mathbf{D}^{-1/2}\mathbf{A}\mathbf{D}^{-1/2}$ is bounded between $[-1,1]$.  As a result,  such a GCNN  model is uniformly stable  (assuming that the graph features are also normalized appropriately, e.g., $\|\mathbf{x}\|_2=1$). \\

\noindent \textbf{Random Walk Graph Filters}: Another graph filter that has been widely used is based on random walks: $g(\mathbf{L}) = \mathbf{D}^{-1}\mathbf{A}  + \mathbf{I}$~\cite{puy2017unifying}. The eigenvalues of $\mathbf{D}^{-1}\mathbf{A} $ are spread out in the interval $[0, 2]$ and thus such a GCNN model is uniformly stable.   \\

\noindent \textbf{Importance of Batch-Normalization in GCNN}: Recall that $ \mathbf{g}_\lambda =\underset{\mathbf{x}}{\sup}  \hspace{0.5em} \Big|\sum_{j \in  \Na(\mathbf{x})}   e_{\cdot j }  \mathbf{x}_{j}  \Big| $    and notice that in Equation~(\ref{eq:g_lambda}), we assume that the graph signals  are normalized  in order to bound $\mathbf{g}_{\lambda}$ . This can easily be accomplished by normalizing features during data pre-processing phase for a single layer GCNN. However, for a multi-layer  GCNN, the  intermediate  feature outputs are  not guaranteed to be normalized. Thus to ensure stability, it is crucial   to employ batch-normalization layers in GCNN models. This   has already been reported in~\cite{xu2018powerful} as an important factor for keeping the  GCNN outputs stable.

\section{Experimental Evaluation}\label{sec:exp_results}
In this section, we empirically  evaluate the effect of graph filters on the GCNN stability bounds using four different GCNN filters. We employ three citation network datasets: Citeseer, Cora and Pubmed (see~\cite{kipf2016semi} for  details about the datasets). \\

\noindent \textbf{Experimental Setup}: We extract $1-$hop ego-graphs of each node in   a given dataset to create samples and normalize the node graph features such that $\|\mathbf{x}\|_2=1$ in the data pre-processing step. We run the SGD algorithm with a fixed learning rate $\eta=1$  with the batch size equal to $1$ for $100$ epochs on all datasets. We employ ELU (set $\alpha=1$) as the activation function   and  cross-entropy as the loss function.   \\

\noindent \textbf{Measuring Generalization Gap}:    In this experiment, we quantitatively measure the generalization gap  defined as the absolute difference between the training and test errors.
From Figure~\ref{fig:gen_gap}, it is clear that the unnormalized graph convolution filters such as $g(\mathbf{L}) = \mathbf{A} + \mathbf{I}$ show  a significantly higher generalization gap than the normalized ones such as $\mathbf{D}^{-1/2}\mathbf{A}\mathbf{D}^{-1/2}$ or random walk $g(\mathbf{L}) = \mathbf{D}^{-1}\mathbf{A}  + \mathbf{I}$ based graph filters. The results hold consistently across the three datasets.  We note that the generalization gap becomes constant  after a certain number of iterations. While this phenomenon is not reflected in our bounds, it can plausibly be explained by considering  the variable bounding parameters (as a function of SGD iterations). This hints at the pessimistic nature of our bounds. \\

\noindent \textbf{Measuring  GCNN  Learned Filter-Parameters Stability Based On SGD Optimizer}:  In this experiment, we evaluate the difference between learned weight parameters of two single layer GCNN models trained on datasets $S$ and $S^{i}$ which differ   precisely in one sample point. We generate $S^{i}$ by surgically altering one sample point in $S$
at the node index $i=0$. For this experiment, we initialize the GCNN models on both datasets with the same parameters and random seeds, and then run the SGD algorithm. After each epoch, we measure the $L2-$norm difference between the   weight parameters of the respective models. From Figure~\ref{fig:parms_norm}, it is evident that   for the unnormalized graph convolution filters,  the weight parameters  tend to deviate by a large amount and therefore the network is less stable. While for the normalized graph filters the norm difference converges quickly to a fixed value.  These empirical observations are reinforced   by our stability bounds. However, the decreasing trend in the norm difference after a certain number of iterations before convergence,  remains unexplained,  due to the pessimistic nature of our bounds.


\section{Conclusion and Future Work}\label{sec:conclusion}

We have taken the  first steps towards establishing a deeper theoretical understanding of GCNN models  by analyzing their stability and  establishing  their generalization guarantees.  More specifically, we have shown that the algorithmic stability  of GCNN models depends upon the largest absolute eigenvalue of graph convolution filters. To ensure uniform stability and thereby generalization guarantees, the largest absolute eigenvalue   must be     independent of the graph size.
Our results shed new insights on the design of 
new \& improved graph convolution filters with guaranteed algorithmic stability. Furthermore, applying our results to existing GCNN models, we provide a theoretical justification for  the importance of employing the batch-normalization process in  a GCNN   architecture. We have also conducted empirical evaluations based on real world datasets which support our theoretical findings. To the best of our knowledge, we are the first to study stability bounds on graph learning in a semi-supervised setting and derive generalization bounds for GCNN  models.

As part of our ongoing and future work, we will extend our analysis to multi-layer GCNN models. For a multi-layer GCNN,  we   need to bound the difference in weights at each layer according to the back-propagation  algorithm. Therefore the main technical challenge  is to study the stability of the full fledged back-propagation algorithm. Furthermore, we plan to study the stability and generalization properties of non-localized convolutional filters designed based on rational polynomials of the graph Laplacian. We also plan to generalize our analysis framework beyond semi-supervised learning to provide generalization guarantees in   learning settings where multiple graphs are present, e.g.,  for graph classification. 

\section{Acknowledgments}

The research was supported in part by US DoD DTRA grant HDTRA1-14-1-0040,  and NSF grants CNS
1618339, CNS 1617729, CNS 1814322 and CNS183677.


\bibliographystyle{ACM-Reference-Format}
\bibliography{refs}

\newpage

\section{Appendix}

\noindent\textbf{Proof of Theorem~\ref{thm:generlization_error}}:  To
derive generalization bounds for  uniform stable randomized
algorithms, we utilize McDiarmid's concentration inequality. Let
$\mathbf{X}$ be a random variable set and $f:\mathbf{X}^{m}
\rightarrow R$, then the inequality is given as,

\begin{equation}
\begin{split}
& if \underset{x_1,..,x_i,..,x_m,x_i^{'}}{\sup}|f(x_1,..,x_i,..,x_m)-f(x_1,..,x_i^{'},..,x_m)| \leq c_i \\
& =\underset{x_1,..,x_i,..,x_m,x_i^{'}}{\sup}|f_S-f_{S^{i}}| \leq c_i \hspace{1em},\forall i \\
& \implies P\Big( f(S)-\mathbf{E}_S[f(S)] \geq \epsilon \Big) \leq e^{-\frac{2 \epsilon^2}{\sum_{i=1}^{m}c_i^2}} \\
\end{split}
\end{equation}

We will derive some expressions that would be helpful to compute variables needed for applying McDiarmid's inequality.

Since the samples are i.i.d., we have

\begin{equation}~\label{lemma:expectation_form}
\begin{split}
\mathbf{E}_{S}[\ell(A_S,\mathbf{z})] &=\int \ell\big(A(\mathbf{z}_1,...,\mathbf{z}_m),\mathbf{z}\big)p(\mathbf{z}_1,...,\mathbf{z}_m) d\mathbf{z}_1...d\mathbf{z}_m \\
&=\int \ell\big(A(\mathbf{z}_1,...,\mathbf{z}_m),\mathbf{z}\big)p(\mathbf{z}_1)...p(\mathbf{z}_m)  d\mathbf{z}_1...d\mathbf{z}_m \\
\end{split}
\end{equation}

Using Equation~\ref{lemma:expectation_form} and   renaming the variables, one can show that

\begin{equation}~\label{lemma:rename_form2}
\begin{split}
&\mathbf{E}_{S}[\ell(A_S,\mathbf{z}_j)] =\int \ell\big(A(\mathbf{z}_1,..,\mathbf{z}_j,..,\mathbf{z}_m),\mathbf{z}_j\big) \times \\
& \hspace{10em} p(\mathbf{z}_1,..,\mathbf{z}_j,..,\mathbf{z}_m) d\mathbf{z}_1...d\mathbf{z}_m \\
& = \int \ell\big(A(\mathbf{z}_1,..,\mathbf{z}_j,..,\mathbf{z}_m),\mathbf{z}_j\big)p(\mathbf{z}_1)..p(\mathbf{z}_j)..p(\mathbf{z}_m) d\mathbf{z}_1...d\mathbf{z}_m \\
& = \int \ell\big(A(\mathbf{z}_1,..,\mathbf{z}_i^{'},..,\mathbf{z}_m),\mathbf{z}_i^{'}\big)p(\mathbf{z}_1)..p(\mathbf{z}_i^{'})..p(\mathbf{z}_m) d\mathbf{z}_1..d\mathbf{z}_i^{'}..d\mathbf{z}_m \\
& = \int \ell\big(A(\mathbf{z}_1,..,\mathbf{z}_i^{'},..,\mathbf{z}_m),\mathbf{z}_i^{'}\big)p(\mathbf{z}_1,..,\mathbf{z}_i^{'},..,\mathbf{z}_m) d\mathbf{z}_1..d\mathbf{z}_i^{'}..d\mathbf{z}_m \times \\
& \hspace{10em}  \int p(\mathbf{z}_i) d\mathbf{z}_i\\
& = \int \ell\big(A(\mathbf{z}_1,..,\mathbf{z}_i^{'},..,\mathbf{z}_m),\mathbf{z}_i^{'}\big)p(\mathbf{z}_1,..,\mathbf{z}_i,\mathbf{z}_i^{'},..,\mathbf{z}_m) d\mathbf{z}_1...d\mathbf{z}_m d\mathbf{z}_i^{'}\\
&=\mathbf{E}_{S,z_{i}^{'}}[\ell(A_{S^{i}},\mathbf{z}_i^{'})]
\end{split}
\end{equation}

Using Equation~\ref{lemma:rename_form2} and  $\beta-$uniform stability, we obtain

\begin{equation}
\begin{split}
&\mathbf{E}_S[\mathbf{E}_{A}[R(A)]-\mathbf{E}_{A}[R_{emp}(A)]] = \mathbf{E}_{S}[\mathbf{E}_z[\mathbf{E}_{A}[\ell(A_S,\mathbf{z})]]]-  \\
& \hspace{10em}\frac{1}{m}\sum\limits_{j=1}^{m}\mathbf{E}_{S}[\mathbf{E}_{A}[\ell(A_S,\mathbf{z}_j)]] \\
&= \mathbf{E}_{S}[\mathbf{E}_z[\mathbf{E}_{A}[\ell(A_S,\mathbf{z})]]]-\mathbf{E}_{S}[\mathbf{E}_{A}[\ell(A_S,\mathbf{z}_j)]] \\
& =\mathbf{E}_{S,z_{i}^{'}}[\mathbf{E}_{A}[\ell(A_S,\mathbf{z}_{i}^{'})]]- \mathbf{E}_{S,z_{i}^{'}}[\mathbf{E}_{A}[\ell(A_{S^{i}},\mathbf{z}_i^{'})]]\\
& =\mathbf{E}_{S,z_{i}^{'}}[\mathbf{E}_{A}[\ell(A_S,\mathbf{z}_{i}^{'})-\ell(A_{S^{i}},\mathbf{z}_{i}^{'})]] \\
& \leq \mathbf{E}_{S,z_{i}^{'}}[\mathbf{E}_{A}[|\ell(A_S,\mathbf{z}_{i}^{'})-\ell(A_{S^{i}},\mathbf{z}_{i}^{'})|]]\\
& \leq 2\beta \\
\end{split}
\end{equation}

%

\begin{equation}~\label{lemma:gen_bound}
\begin{split}
& |\mathbf{E}_{A}[R(A_S)-R(A_{S^{ i}})]|=  |\mathbf{E}_z[\mathbf{E}_{A}[\ell(A_S,\mathbf{z})]]-\mathbf{E}_z[\mathbf{E}_{A}[\ell(A_{S^{ i}},\mathbf{z})]]| \\
& =	|\mathbf{E}_z[\mathbf{E}_{A}[\ell(A_S,\mathbf{z})] - \mathbf{E}_{A}[\ell(A_{S^{ i}},\mathbf{z})]] |\\
& \leq \mathbf{E}_z[\mathbf{E}_{A}[|\ell(A_S,\mathbf{z})]-\mathbf{E}_{A}[\ell(A_{S^{ i}},\mathbf{z})|]] \\
& \leq \mathbf{E}_z[\beta]=2\beta \\
\end{split}
\end{equation} 

%

\begin{equation}~\label{lemma:emp_bound}
\begin{split}
& |\mathbf{E}_{A}[R_{emp}(A_S)]- R_{emp}(A_{S^{i}})]|   \leq \\ & \hspace{8em} |\frac{1}{m}\sum\limits_{j=1,j \neq i}^{m} (\mathbf{E}_{A}[\ell(A_S,\mathbf{z}_j)-\ell(A_{S^{i}},\mathbf{z}_j)])|  + \\ & \hspace{8em} |\frac{1}{m} (\mathbf{E}_{A}[\ell(A_S,\mathbf{z}_i)-\ell(A_{S^{i}},\mathbf{z}_i^{'})]) |\\
& \leq  2\frac{(m-1)}{m}2\beta +\frac{M}{m} \\
& \leq   2\beta +\frac{M}{m} \\
\end{split}
\end{equation}


Let $K_{S}:=R(A_{S})-R_{emp}(A_{S})$.

Using Equation~\ref{lemma:gen_bound} and Equation~\ref{lemma:emp_bound}, we have

\begin{equation}
\begin{split}
& |\mathbf{E}_A[K_{S}]-\mathbf{E}_A[K_{S^{i}}]| =\Big| \mathbf{E}_A[\big(R(A_{S})-R_{emp}(A_{S}) \big)] \\
& \hspace{10em} - \mathbf{E}_A[\big( R(A_{S^{i}}) - R_{emp}(A_{S^{i}})\big)] \Big| \\
& \leq \Big| \mathbf{E}_A[R(A_{S})]-\mathbf{E}_A[R(A_{S^{i}})] \Big| + \Big| \mathbf{E}_A[R_{emp}(A_{S})] \\
& \hspace{14em}-\mathbf{E}_A[R_{emp}(A_{S^{i}})] \Big|\\
& \leq 2\beta+(2\beta+\frac{M}{m}) \\
& \leq 4\beta+\frac{M}{m} \\
\end{split}
\end{equation} 

Applying McDiarmid's concentration inequality, 

$$P \Bigg(  \mathbf{E}_A[K_{S}]-\mathbf{E}_S[\mathbf{E}_A[K_{S}]] \geq \epsilon   \Bigg) \leq \underbrace{e^{-\frac{2\epsilon^2}{m(4\beta+\frac{M}{m})^2}}}_{\delta}$$

$$P \Bigg(  \mathbf{E}_A[K_{S}] \leq  2\beta +  (4m\beta+M)\sqrt{\frac{\log \frac{1}{\delta}}{2m}} \Bigg) \geq 1-\delta$$

This complete the proof of Theorem~\ref{thm:generlization_error}.

\end{document}